\documentclass{article}

\usepackage[nonatbib, main, final]{neurips_2026}

\usepackage[utf8]{inputenc} 
\usepackage[T1]{fontenc}    
\usepackage{hyperref}       
\usepackage{url}            
\usepackage{booktabs}       
\usepackage{amsfonts}       
\usepackage{nicefrac}       
\usepackage{microtype}      
\usepackage{xcolor}         
\usepackage{graphicx}
\usepackage{amsmath}
\usepackage{pifont}
\usepackage{fontawesome5}
\usepackage{subcaption}
\usepackage{enumitem}
\usepackage{makecell}
\usepackage{multirow}
\hypersetup{hidelinks}

\title{Emergence of a Shared Canonical Object Frame from In-the-Wild Videos} 

\author{%
  Tom Fischer$^{1}$ \quad
  Martin Sundermeyer$^{2}$ \quad
  Adam Kortylewski$^{3}$ \quad
  Eddy Ilg$^{1,\dagger}$
  \\
  $^{1}$University of Technology Nuremberg \\
  $^{2}$Google \\
  $^{3}$CISPA Helmholtz Center for Information Security
}

\newcommand{\Sheep}[1][]{\includegraphics[width=10pt,trim={6cm 7cm 5cm 6cm},clip]{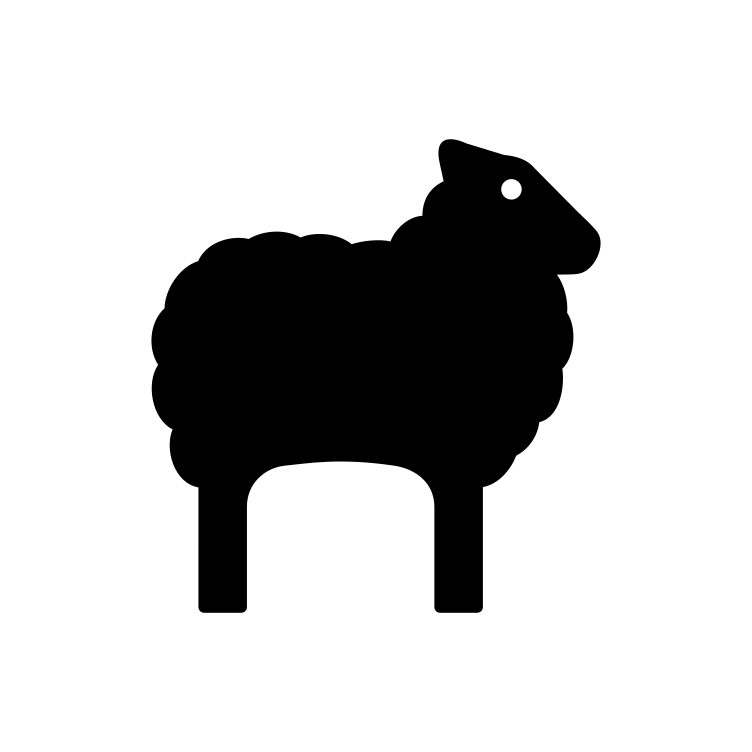}}
\newcommand{\Backpack}[1][]{\includegraphics[height=10pt]{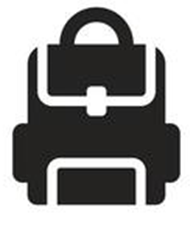}}
\newcommand{\Bench}[1][]{\includegraphics[height=10pt]{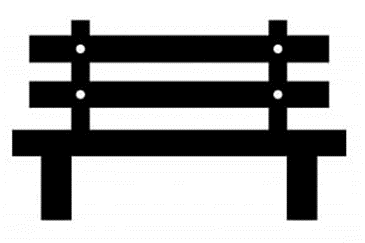}}
\newcommand{\Hairdryer}[1][]{\includegraphics[height=10pt]{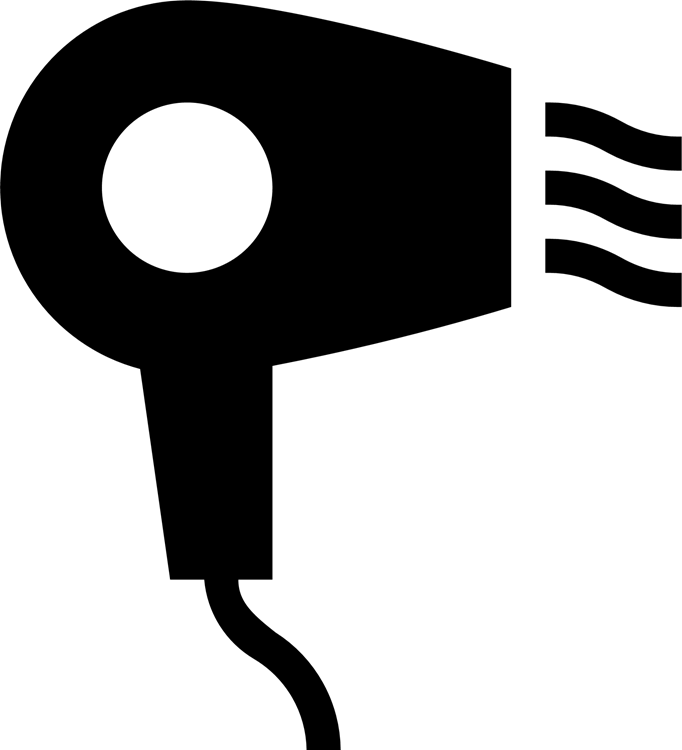}}
\newcommand{\Microwave}[1][]{\includegraphics[height=10pt]{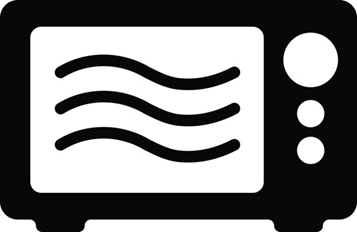}}
\newcommand{\Mouse}[1][]{\includegraphics[height=10pt]{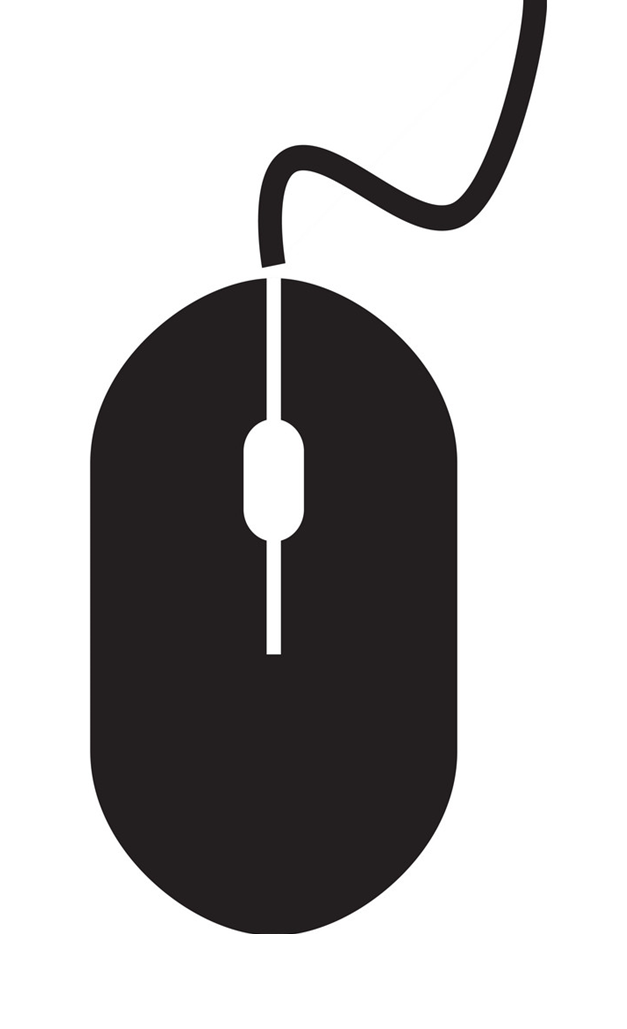}}
\newcommand{\Remote}[1][]{\includegraphics[height=10pt]{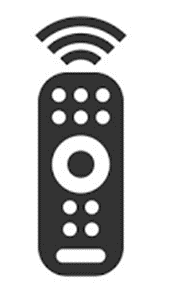}}
\newcommand{\Toaster}[1][]{\includegraphics[height=10pt]{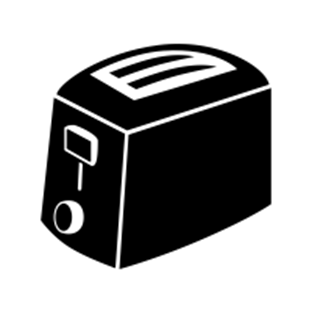}}
\newcommand{\Toilet}[1][]{\includegraphics[height=10pt]{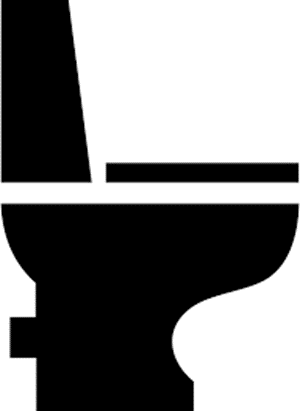}}

\DeclareMathOperator*{\argmin}{arg\,min}

\begin{document}

\maketitle
{\renewcommand{\thefootnote}{$\dagger$}\footnotetext{Now at Google; work done while at the University of Technology Nuremberg.}}

\begin{figure}[h]
\centering
\includegraphics[width=\linewidth]{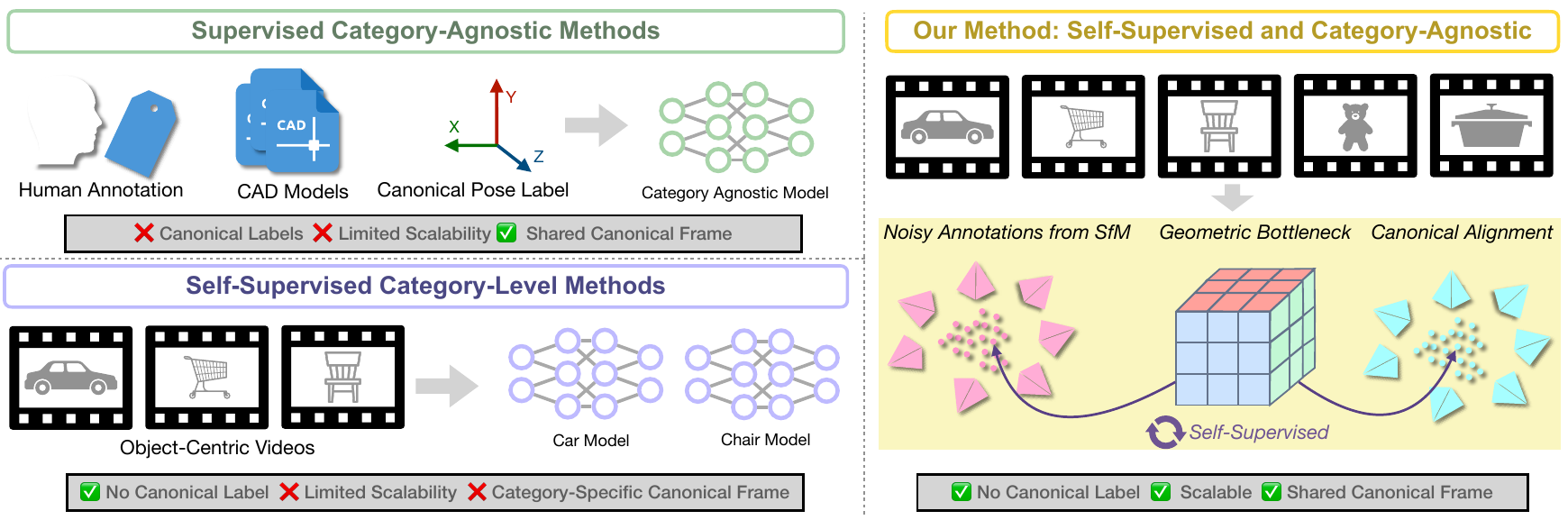}
\caption{\textbf{Three approaches for learning of canonical object frames.}
\emph{Top left:} Supervised methods can obtain a shared canonical frame from rendering aligned CAD models or manual pose annotation, which is expensive to scale.
\emph{Bottom left:} Self-supervised category-level methods learn from object-centric videos without canonical labels, but train separate models per category, limiting scalability and producing no shared frame across categories.
\emph{Right:} Our approach trains a single model across all categories by routing noisy SfM reconstructions through a shared geometric bottleneck, producing a shared canonical frame without canonical pose labels or category conditioning.}
\label{fig:teaser}
\end{figure}

\begin{abstract}
Comparing object orientations and positions across different instances requires their poses to be expressed in a shared canonical frame. 
Establishing such frames has traditionally required manual annotation, creating a scaling bottleneck that limits category and instance diversity.
We show that a shared canonical frame can instead emerge from self-supervised training on object-centric videos of rigid objects captured in the wild, using only noisy camera poses from Structure-from-Motion.
Our key idea is to route all training sequences through a shared geometric bottleneck: a coarse canonical mesh that carries no category-specific detail.
By learning dense correspondences from image pixels to this mesh, and estimating per-sequence alignments from noisy SfM geometry, a common canonical frame emerges from multi-view consistency and the semantic priors of the feature extractor, without any canonical pose labels or category conditioning.
Trained in a self-supervised manner on ${\sim}160{,}000$ in-the-wild object videos, our method achieves competitive accuracy on category-level pose estimation benchmarks compared to methods that rely on canonical pose supervision. The code and checkpoint are available on \href{https://github.com/Fischer-Tom/Emergent-Canonical-Frame/}{github.com/Fischer-Tom/Emergent-Canonical-Frame/}.
\end{abstract}

\section{Introduction}

A shared canonical frame enables consistent reasoning about object orientation across different instances. 
This is essential in applications like augmented reality~\cite{surveyAR}, robotic manipulation~\cite{simeonov2022neural, pan2025omnimanip}, category-level pose estimation~\cite{nocs}, shape retrieval~\cite{tangelder2004survey}, and controlled 3D generation~\cite{lu2025orientation, parihar2025compass}. 
However, establishing canonical frames has traditionally required substantial manual supervision, either via aligned CAD models~\cite{chang2015shapenet, deitke2023objaverse}, controlled rendering of synthetic assets~\cite{omni6d,ov9d}, or careful real-world pose annotation~\cite{nocs, pascal3d, imagenet3d}. 
This reliance on canonical supervision creates a scaling bottleneck: synthetic data introduces a domain gap, while annotated real data remains limited in category and instance diversity. 
Learning canonical object representations that generalize broadly across real-world data therefore remains an open challenge.

A natural question is whether canonical frames can emerge from self-supervised training on readily available data, following trends in representation learning where scale and data diversity drive generalization~\cite{dino, clip}. 
Object-centric videos of rigid objects are abundant, and Structure-from-Motion (SfM) can recover camera poses and coarse geometry from them without manual annotation. 
However, SfM reconstructions are sequence-specific: each video yields internally consistent poses, but in an arbitrary coordinate system. 
Two videos of different cars may each produce accurate relative poses, yet their coordinate frames bear no relation to each other or to any shared notion of canonical orientation. 
Geometric registration (e.g., ICP) can align sufficiently similar shapes, but cannot establish shared notions of front, up, or left across objects with different geometry or topology.
Learning a shared frame therefore requires a signal that transfers semantic orientation across sequences, which per-sequence geometry alone does not provide.

In this work, we show that a shared canonical frame can emerge from object-centric videos without canonical pose labels (see Figure~\ref{fig:teaser}). 
Our approach routes all training sequences through a shared geometric bottleneck: a coarse canonical mesh defined in a fixed coordinate frame. 
A correspondence network learns to map image pixels to vertices of this mesh, while estimated per-sequence alignments transform noisy SfM camera poses into the mesh's frame before each loss computation. 
SfM thus enforces multi-view consistency within each sequence, while the shared correspondence network transfers semantic orientation cues across sequences through visual appearance.
Because the canonical mesh carries no category-specific detail, similar object parts across different instances consistently map to the same surface regions, and a common canonical frame emerges from multi-view consistency and the semantic priors of the feature extractor.
At inference, our model takes a single RGB image and predicts the object's 3D rotation and scale-normalized translation in a canonical frame shared across all categories.

We demonstrate the utility of the learned correspondences on category-level pose estimation, achieving accuracy competitive with methods trained using canonical pose labels across multiple benchmarks. 
Our training uses ${\sim}$160k in-the-wild object-centric videos without category conditioning or canonical pose annotations, enabling generalization to object categories not seen during training (Section~\ref{sec:exp_analysis}). 
In summary, our contributions are as follows:

\begin{itemize}[leftmargin=0.5cm]
    \item \textbf{Self-supervised canonical frame emergence.} We show that a shared canonical frame can emerge from object-centric videos using only noisy SfM camera poses, without any canonical pose annotation.
    \item \textbf{Geometric bottleneck.} We show that routing all training sequences through a shared, coarse canonical mesh induces cross-sequence consistency and enables a common reference frame to emerge through a combination of multi-view alignment and strong pre-trained features.
    \item \textbf{Category-agnostic generalization.} Without canonical pose supervision, our self-supervised approach achieves competitive performance with canonically supervised methods across multiple pose estimation benchmarks, including categories unseen during training.
\end{itemize}

\begin{figure}[t]
    \centering
    \includegraphics[width=\linewidth]{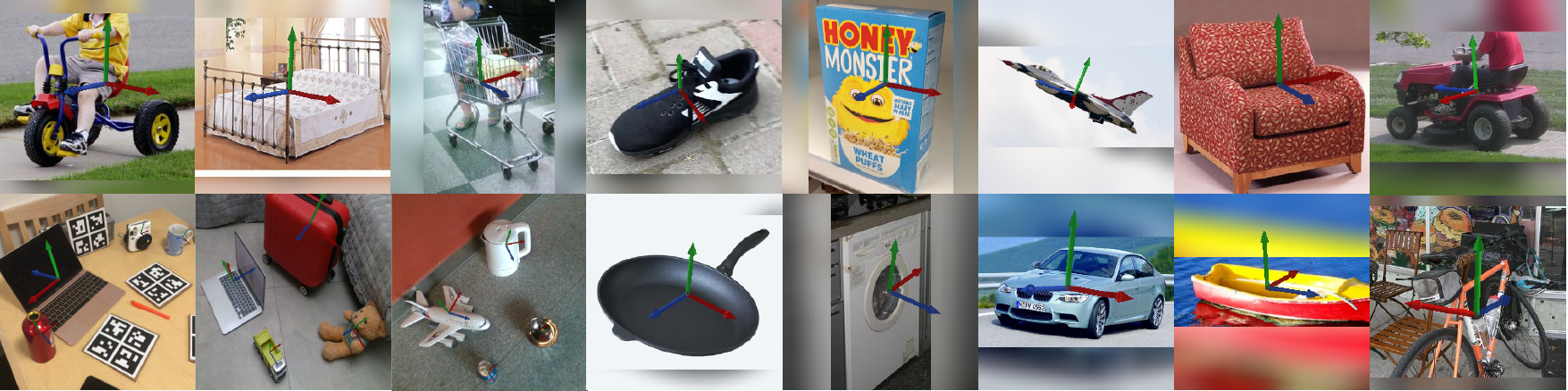}
\caption{\textbf{Qualitative pose predictions.} 
Predicted poses (3D rotation and scale-normalized translation) visualized as canonical axes overlaid on images from diverse categories and benchmarks. 
A single fixed rotation is applied uniformly to all predictions for visualization, without per-category or per-dataset adjustments. 
Semantic directions such as ``front'' and ``up'' of object instances are not explicitly supervised, but tend to consistently map onto the same axes in the learned canonical frame.
We also show limitations discussed in Section~\ref{sec:concl}: the laptop is articulated, so its predicted frame follows the dominant part; left and right shoes share one canonical frame, since chirality is not modeled.
}
    \label{fig:qualitative_main}
\end{figure}

\section{Related Work}
\label{sec:related}

\textbf{Canonical frames for pose supervision.} 
Category-level pose estimation typically assumes a shared canonical object frame across instances. Wang~\emph{et al.}~\cite{nocs} introduced NOCS, which predicts normalized correspondences and recovers pose via Umeyama~\cite{umeyama}. 
Subsequent work improves accuracy using RGB-D cues such as shape priors and deformation models~\cite{sgpa,gpvpose,dpdn}, or stronger architectures~\cite{istnet,secondpose}. 
RGB-only methods~\cite{oldnet,msos,dmsr,unipose,lapose} operate on appearance alone, enabling shape-agnostic representations at the cost of metric accuracy. 
Neural mesh models~\cite{nemo3d,inemo,novum} attach learned features to a geometric scaffold and recover pose by aligning rendered features to image observations, with extensions to full 6D and joint detection-and-pose~\cite{nemo6d,unipose}. 
Recent open-world orientation estimation methods~\cite{orientanything,orientanythingv2} demonstrate impressive generalization across diverse categories, but still rely on canonical supervision through large-scale synthetic data with predefined canonical frames.
Across these approaches, the dependence on canonical pose labels creates a scaling bottleneck: real-world datasets remain limited in category and instance diversity~\cite{nocs,wild6d,housecat6d,pascal3d,imagenet3d}, and synthetic pipelines inherit a domain gap. 
We show that canonical frames can instead emerge from category-agnostic training on noisy in-the-wild data, without canonical pose labels or synthetic samples.

\textbf{Novel object pose estimation without canonical frames.} 
A separate line of work addresses pose estimation for objects unseen during training, but relies on additional object information at test time. 
Onboarding-based methods use CAD renderings~\cite{ornek2024foundpose,gigapose,megapose} or multi-view captures~\cite{wen2024foundationpose,sun2022onepose,he2022onepose++,gen6d} to build object-specific templates or reconstructions, and then estimate pose by matching test images to these references. 
Reference-based methods instead predict relative transformations between a test image and a reference view, either directly~\cite{liu2025unopose,liu2025one2any,kuang2025conceptpose} or via single-view 3D reconstructions~\cite{lee2025any6d,geng2025one}. 
Recent methods such as MFOS~\cite{lee2024mfos} and
BoxDreamer~\cite{yu2025boxdreamer} also use a cuboid primitive for generalizable pose estimation, but work in the reference
view setting rather than recovering a shared canonical frame.
Both families avoid canonical supervision, but the resulting coordinate frame is instance- or reference-defined: two different chairs yield unrelated orientations that are not directly comparable across objects. 
We instead seek a shared canonical frame consistent across instances and categories, without canonical labels or test-time onboarding.

\textbf{Emergence of canonical frames from object-centric videos.} 
The most closely related line of work reduces supervision by exploiting multi-view geometry from object-centric videos. 
SHIC~\cite{shtedritski2024shic} predicts 2D-3D correspondences by conditioning features on detailed templates.
UOP3D~\cite{sommer2024unsupervised} shows that such templates can instead be reconstructed from video, removing the need for manual template design, and Common3D~\cite{sommer2025common3d} extends this to category-specific morphable models learned entirely from object-centric sequences. 
While these methods progressively reduce supervision, they still require category labels and train separate models per category, introducing a scaling bottleneck as the number of categories grows. 
Canonical shape learning~\cite{li2021leveraging,sun2021canonical}
shows that consistent frames can emerge without supervision on 3D
point clouds, but does not extend to RGB. Multi-Path
Encoders~\cite{sundermeyer2020multi} learn shared latent spaces from
RGB without pose supervision, but scale linearly with object count.

Self-supervised correspondence learning has shown that vision foundation features can produce semantically meaningful correspondences without manual keypoint annotation~\cite{zhang2023tale}, and recent work couples features to geometric primitives such as spheres to reduce ambiguities~\cite{mariotti2024improving,dunkel2025doit}.
While these methods share our use of a geometric bottleneck, they target pairwise keypoint matching rather than canonical pose recovery, which additionally requires a globally coherent 3D frame.

\textbf{Summary.} Existing approaches either (a) assume canonical frames as supervision targets, (b) adopt instance-specific reference frames or test-time onboarding, or (c) use self-supervised training to recover canonical frames, but only within category-specific pipelines. 
Our work addresses the gap among these directions: a shared canonical frame emerges from self-supervised training on in-the-wild object-centric video through a geometric bottleneck, enabling category-agnostic pose estimation that transfers to unseen object categories.

\begin{figure}[t]
\centering
\includegraphics[width=\linewidth]{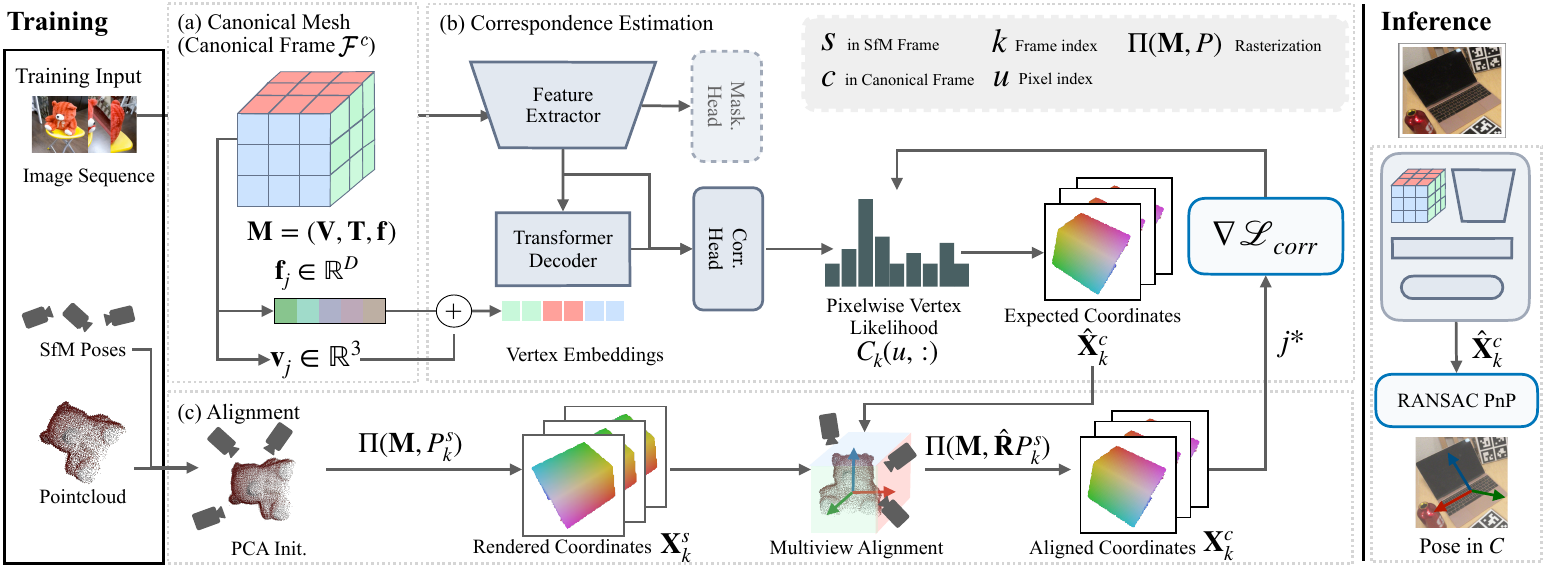}
\caption{\textbf{Method overview.}
(a) We define the shared geometric bottleneck as a canonical mesh in a fixed coordinate frame.
(b) Each image is processed independently: the correspondence network predicts dense 2D--3D correspondences from pixels to mesh vertices.
(c) During training, we estimate a per-sequence rotation that aligns each sequence's SfM frame to the canonical mesh frame. This alignment transforms SfM-derived correspondences into the canonical frame, producing pseudo-labels that supervise the correspondence network.
During inference, correspondences are predicted from a single view, and the 3D rotation and scale-normalized translation are recovered via PnP.}
\label{fig:method}
\end{figure}

\section{Method}
\label{sec:method}

Our method learns dense correspondences from image pixels to the surface of a shared canonical mesh. 
Because this mesh is fixed and shared across all training sequences regardless of object category, it acts as a geometric bottleneck: consistent correspondences give rise to a shared canonical frame from which object pose can be recovered.
We assume rigid objects throughout; articulated and deformable objects are outside the scope of this work.

Training relies on object-centric video sequences with SfM camera poses and off-the-shelf segmentation masks. 
No canonical pose annotations or category labels are used to condition our model. 
The method has three components: (a) the canonical mesh that defines the coordinate frame, (b) a correspondence network that maps image pixels to vertices of this mesh, and (c) a per-sequence alignment step that transforms SfM camera poses into the mesh's frame before loss computation.
An overview is shown in Figure~\ref{fig:method}.

\subsection{Canonical Mesh}
\label{sec:mesh}

The canonical mesh $\mathbf{M}=(\mathbf{V}, \mathbf{T}, \mathbf{f})$ is a coarse 3D shape with $M$ vertices uniformly sampled on its surface and a set of triangular faces $\mathbf{T}$ (see Figure~\ref{fig:method} (a)).
Each vertex $\mathbf{v}_j\in\mathbf{V}$ carries a learnable embedding $\mathbf{f}_j\in\mathbb{R}^D$ that serves as a query in the correspondence network (Section~\ref{sec:correspondence}).
The mesh defines a fixed canonical frame $\mathcal{F}^c$: once the correspondence network maps object pixels to mesh vertices consistently, an object's orientation can be recovered from the resulting 2D-3D correspondences.

To generate supervision for the correspondence network during training, we need to know which 3D point on the mesh surface is visible at each pixel under a given camera pose.
We obtain this by rasterizing the mesh under camera pose $P$ (denoted $\Pi(\mathbf{M}, P)$) and interpolating vertex positions via barycentric coordinates over the mesh faces.
This produces a dense coordinate map $\mathbf{X} \in \mathbb{R}^{N \times 3}$, where each visible pixel stores a 3D point on the mesh surface, along with a binary visibility mask.

The canonical mesh is a coordinate scaffold, not a geometric model of the object.
It carries no category- or instance-specific geometric detail, so that the correspondence network must rely on appearance rather than shape matching.
Consequently, the object-to-mesh mapping need not preserve topology or be bijective: a tire, for example, is mapped onto the mesh as a whole, and its hole is simply treated as part of the object.
We use a unit cube as the default.
Its faces, edges, and corners provide discrete orientation anchors, while its equal side lengths leave the three axes interchangeable, so that semantic axes must be assigned from visual evidence rather than from object aspect ratio.
We compare against alternative primitives in Section~\ref{sec:ablations}.

\subsection{Correspondence Prediction}
\label{sec:correspondence}

Given an input image $\mathbf{I}$, the correspondence network predicts, for each pixel, a distribution over the $M$ vertices of the canonical mesh (see Figure~\ref{fig:method} (b)).
From these distributions, we obtain dense 2D-3D correspondences between image pixels and 3D locations on the mesh surface.

A feature extractor produces dense features $\mathbf{F} \in \mathbb{R}^{N \times D}$, where $N$ is the number of pixels and $D$ is the feature dimension.
To obtain per-vertex features, we use a transformer decoder that takes the pixel features $\mathbf{F}$ as keys and values, and learnable vertex embeddings $\mathbf{f} \in \mathbb{R}^{M \times D}$ as queries.
Vertex positions $\mathbf{V} \in \mathbb{R}^{M \times 3}$ provide positional encoding for the queries, grounding each to a specific location on the canonical mesh.
Given the decoded vertex features $\tilde{\mathbf{f}} \in \mathbb{R}^{M \times D}$, a soft correspondence matrix is computed as:
\begin{equation}
\label{eq:C}
    \mathbf{C} = \mathrm{softmax}\left( \frac{(\mathbf{F} \mathbf{W}_F)(\tilde{\mathbf{f}} \mathbf{W}_f)^\top}{\tau} \right) \in \mathbb{R}^{N \times M},
\end{equation}
where $\mathbf{W}_F, \mathbf{W}_f \in \mathbb{R}^{D \times D'}$ are learned projections and $\tau$ is a temperature parameter.
Each row $\mathbf{C}(u,:)$ is a distribution over mesh vertices for pixel $u$, and the expected canonical 3D coordinate is:
\begin{equation}
    \hat{\mathbf{x}}^c_u = \sum_{j=1}^{M} \mathbf{C}(u,j)\,\mathbf{v}_j
    \quad \text{or} \quad
    \hat{\mathbf{X}}^c = \mathbf{C}\mathbf{V} \in \mathbb{R}^{N \times 3}.
\end{equation}
We frame correspondence as classification over discrete vertices rather than regression of continuous 3D coordinates; under a softmax, the cross-entropy objective concentrates probability on the target vertex while suppressing all others, encouraging discriminative correspondences.

\subsection{Per-Sequence Alignment}
\label{sec:alignment}

Each SfM reconstruction lives in an arbitrary coordinate frame $\mathcal{F}^s$.
To supervise the correspondence network, we need to express SfM camera poses in the canonical mesh frame $\mathcal{F}^c$.
This requires estimating, for each training sequence, a rotation $\hat{\mathbf{R}}$ that maps $\mathcal{F}^s$ to $\mathcal{F}^c$.
The process is illustrated in Figure~\ref{fig:method} (c) and its effect on the pseudo-labels in Figure~\ref{fig:supp_alignment} (appendix).

We initialize each sequence's alignment via PCA on the masked SfM point cloud, which assigns consistent axes based on geometric extent (e.g., elongated objects share a dominant axis). 
This geometric prior captures coarse canonical structure but cannot resolve semantic orientation (e.g., front-vs-back) and fine-grained alignment.
The per-sequence alignment refines this initialization during training using the correspondence network's predictions.

Concretely, for each of $K$ sampled views, every pixel where the canonical mesh is visible provides a 3D--3D correspondence pair: one point from rasterizing the mesh under the SfM pose $P_k^s$ (in $\mathcal{F}^s$), and one from the network's dense prediction $\hat{\mathbf{X}}_k^c = \mathbf{C}\mathbf{V}$ (in $\mathcal{F}^c$).
Since all views in a sequence share the same SfM frame, we estimate a single rotation from these pairs jointly across views. 
Since the mesh is centered around the origin and at the same scale, we normalize each 3D vector to unit length and solve:
\begin{equation}
\label{eq:wahba}
    \hat{\mathbf{R}} = \argmin_{\mathbf{R} \in \mathrm{SO}(3)} \sum_{k=1}^{K} \sum_{u=1}^{N_k} w_{ku} \left\| \hat{\mathbf{x}}^c_{k,u} - \mathbf{R} \, \mathbf{x}^s_{k,u} \right\|^2,
\end{equation}
where $w_{ku} = H(\mathbf{C}_k(u,:))^{-1}$ with $H(\mathbf{p})=-\sum_j p_j\log p_j$ is the inverse entropy of the correspondence distribution at pixel $u$, so that confident predictions contribute more to the alignment.
We add a small constant ($\epsilon=10^{-8}$) for numerical stability in the entropy computation and normalize weights within each view to unit mean. 
This is Wahba's problem~\cite{wahba} and admits a closed-form solution via SVD, which we wrap in a standard RANSAC loop to reject outlier correspondences.

The estimated rotation transforms the SfM poses into the canonical frame: $P^{c}_k = \hat{\mathbf{R}} P^s_k$. 
These aligned poses are then used to rasterize the canonical mesh and generate pseudo-labels for the correspondence loss, as described in Section~\ref{sec:training}.
Since the alignment depends on the network's current predictions, and the resulting pseudo-labels in turn update the network, both improve together during training.

\subsection{Training Objective}
\label{sec:training}

Once the SfM poses have been transformed into the canonical frame, we use them to generate pseudo-labels for the correspondence network.
For each view $k$, we rasterize the canonical mesh under the aligned pose $P_k^c$, producing a canonical 3D coordinate $\mathbf{x}_{k,u}^c$ for each visible pixel $u$.
We assign each coordinate to its nearest mesh vertex:
\begin{equation}
j^{*}_{k,u} = \argmin_{j} \|\mathbf{x}^{c}_{k,u} - \mathbf{v}_j\|_2,
\end{equation}
and supervise the correspondence distribution with cross-entropy:
\begin{equation}
    \mathcal{L}_{\mathrm{corr}} =  \sum_{k=1}^{K} \sum_{u=1}^{N_k} -\log \mathbf{C}_k(u, j^*_{k,u}).
\end{equation}
This encourages the network to predict a distribution that concentrates on the mesh vertex visible at that location under the current alignment.

We additionally train a mask prediction head to identify which pixels belong to the object. 
During training, this head is supervised by the binary visibility mask obtained from rasterizing the canonical mesh under the estimated alignment (Section~\ref{sec:alignment}). 
At inference, the predicted mask determines the object region from which pose is recovered.
The mask head is supervised using binary cross-entropy and Dice loss against the visibility mask $\mathbf{m}^{c}$ obtained from the same rasterization:
\begin{equation}
    \mathcal{L}_{\mathrm{mask}} = \mathrm{BCE}(\hat{\mathbf{m}}, \mathbf{m}^{c}) + \mathrm{Dice}(\hat{\mathbf{m}}, \mathbf{m}^{c}),
\end{equation}
where $\hat{\mathbf{m}}$ is the predicted mask (boundary upweighting details in appendix).
The overall training objective is:
\begin{equation}
    \mathcal{L} = \mathcal{L}_{\mathrm{corr}} + \lambda \mathcal{L}_{\mathrm{mask}},
\end{equation}
where $\lambda$ controls the relative contribution of the mask loss. 

\subsection{Inference}
\label{sec:inference}

At test time, the correspondence network predicts $\mathbf{C}$ and the mask head identifies object pixels in a single forward pass.
From the 2D-3D correspondences $\hat{\mathbf{X}}^c = \mathbf{C}\mathbf{V}$, we recover the object's rotation and translation via EPnP~\cite{epnp} with RANSAC. 
No per-sequence alignment or PCA is involved at inference, since the correspondences directly yield a continuous pose in the canonical frame.
Since metric scale is ambiguous from a single RGB image, translation is estimated only up to a global scale factor, matching the convention of prior RGB-only methods~\cite{lapose, givepose, unipose}.
We therefore refer to the output as 3D rotation and scale-normalized translation rather than metric 6D pose.

\begin{table}[t]
\centering
\small
\setlength\tabcolsep{2pt}
\resizebox{\linewidth}{!}{
\begin{tabular}{l c c | cc | cc | cc | cc | cc | cc}
\toprule
Model & Can. & Train
& \multicolumn{2}{c|}{\textbf{REAL275}}
& \multicolumn{2}{c|}{\textbf{Omni6DPose}}
& \multicolumn{2}{c|}{\textbf{Objectron}}
& \multicolumn{2}{c|}{\textbf{PASCAL3D+}}
& \multicolumn{2}{c|}{\textbf{ImageNet3D}} 
& \multicolumn{2}{c}{\textbf{Avg.}}
\\
\cmidrule(lr){4-5}
\cmidrule(lr){6-7}
\cmidrule(lr){8-9}
\cmidrule(lr){10-11}
\cmidrule(lr){12-13}
\cmidrule(lr){14-15}
& & &
Med$\downarrow$ & Acc30$\uparrow$
& Med$\downarrow$ & Acc30$\uparrow$
& Med$\downarrow$ & Acc30$\uparrow$
& Med$\downarrow$ & Acc30$\uparrow$
& Med$\downarrow$ & Acc30$\uparrow$
& Med$\downarrow$ & Acc30$\uparrow$ \\
\midrule
QWEN3-VL\cite{qwen3} & \ding{52} & R.+S.
& 38.7 & 37.6
& 70.6 & 18.8
& 49.0 & 31.1
& 61.1 & 27.0
& 66.5 & 24.0 
& 57.2 & 27.7
\\
\midrule
OriAny.V1\cite{orientanything} & \ding{52} & S.
& 28.4 & 52.2
& 62.8 & \underline{36.7}
& 18.4 & 60.4
& 18.1 & 71.0
& 29.7 & 50.3 
& 31.5 & 54.1
\\
OriAny.V1$^\dagger$\cite{orientanything} & \ding{52} & R.+S.
& 26.7 & 54.1
& 54.5 & 31.5
& \textbf{15.1} & \underline{67.7}
& \textcolor{lightgray}{\textbf{15.7}} & \textcolor{lightgray}{\underline{78.4}}
& \textcolor{lightgray}{\underline{25.7}} & \textcolor{lightgray}{\textbf{55.6}} 
& 27.5 & 57.5
\\
OriAny.V2$^\dagger$\cite{orientanythingv2} & \ding{52} & R.+S.
& \textbf{21.3} & \underline{57.0}
& \textbf{47.7} & \textbf{39.2}
& 19.8 & 66.2
& \textcolor{lightgray}{\underline{17.5}} & \textcolor{lightgray}{76.5}
& \textcolor{lightgray}{28.1} & \textcolor{lightgray}{54.1} 
& \underline{26.9} & \underline{58.6}
\\
\midrule
Ours & \ding{56} & R.
& \underline{21.8} & \textbf{70.0}
& \underline{49.2} & 35.2
& \underline{15.3} & \textbf{69.2}
& \textbf{15.7} & \textbf{79.8}
& \textbf{25.5} & \underline{55.0} 
& \textbf{25.5} & \textbf{61.8}
\\
\bottomrule
\end{tabular}}
\caption{\textbf{Transfer across benchmarks.}
Median rotation error and Acc$30^\circ$ on REAL275, Omni6DPose, Objectron, PASCAL3D+, and ImageNet3D.
Avg.\ reports the macro-average over all five datasets.
All methods are evaluated with the same symmetry-aware protocol and convention mapping discussed in Section~\ref{sec:exp_setup}.
Despite not using canonical pose labels and training only on real video, our method achieves the best all-dataset average.
Methods with a dagger ($\dagger$) were trained on ImageNet3D (overlapping with PASCAL3D+) and are therefore given an unfair advantage on the gray metrics.}
\label{tab:zeroshot_combined}
\end{table}

\section{Experiments}
\label{sec:experiments}

We evaluate our method on category-level pose estimation benchmarks, studying transfer performance, the effect of data scale, the role of design choices, and properties of the learned frame. 
Unless stated otherwise, we report rotation metrics since translation is recovered only up to scale and most baselines predict rotation only.

\subsection{Experimental Setup}
\label{sec:exp_setup}

\textbf{Training.} We train on UCO3D~\cite{uco3d}, which provides ${\sim}$160k object-centric videos with SfM camera poses, sparse point clouds, segmentation masks, and optional PCA-based coordinate alignments. 
From each sequence, we use the masked point cloud (centered at the origin and normalized to unit scale), the segmentation masks for cropping, and the SfM camera poses. 
We optionally use the provided PCA alignment as initialization (Section~\ref{sec:alignment}). 
We do not use canonical pose annotations and train a single model across all categories without category conditioning.

\textbf{Evaluation.} We evaluate on REAL275~\cite{nocs}, Omni6DPose~\cite{omni6dpose}, PASCAL3D+~\cite{pascal3d}, ImageNet3D~\cite{imagenet3d}, and Objectron~\cite{objectron}. 
For category-constrained self-supervised comparisons, we use PASCAL3D+ and ObjectNet3D~\cite{objectnet3d}.
All methods assume single-object images, so for multi-object datasets we predict per-object pose from crops of the dataset-provided bounding boxes.

\textbf{Baselines.}
For category-agnostic rotation estimation, we compare against
OrientAnything V1/V2~\cite{orientanything,orientanythingv2} and
QWEN3-VL~\cite{qwen3}, all trained with canonical pose supervision
(QWEN3-VL prompt details in appendix). 
For category-specific self-supervised comparisons, we use ZSP~\cite{goodwin2022zero}, UOP3D~\cite{sommer2024unsupervised}, and Common3D~\cite{sommer2025common3d}, which train separate models per category.

We report median geodesic rotation error (deg) and $Acc@30^\circ$ with symmetry-aware evaluation following~\cite{orientanythingv2}. 
Symmetry class labels for each category are derived automatically using an LLM (details in appendix), used only at evaluation time and are applied uniformly across all methods.
In result tables, \emph{Can.}\ indicates whether canonical pose annotations are used during training (\ding{52}/\ding{56}), and \emph{Train} indicates whether the training data was real (R), synthetic (S), or mixed (R+S). 

Our canonical frame can differ from a dataset's annotation convention by a fixed rotation. 
Recovering a particular benchmark convention is not our objective, since such conventions are arbitrary and can differ across datasets per category.
For evaluation, we estimate this mapping per category via rotation averaging~\cite{rotaverage} on held-out data from the training split, never on test data, and apply it uniformly to all test predictions.
In practice, this mapping is largely consistent across categories (see Figure~\ref{fig:canonicalization}), and it reduces to selecting among discrete axis conventions (Section~\ref{sec:exp_analysis}). 
Since such a post-hoc alignment is required whenever a method's convention differs from the dataset's, we apply the same mapping protocol to all baselines for a fair comparison.

\textbf{Implementation details.} We describe all implementational details including architecture, hyperparameters, and training schedule in the appendix.


\begin{table}[t]
\centering
\small
\setlength\tabcolsep{4.5pt}
\renewcommand{\arraystretch}{1.15}
\begin{tabular}{l l c c c c}
\toprule
Model
& Method 
& Training Videos
& PASCAL3D+
& ObjectNet3D
& Mean \\
\midrule
\multirow{3}{*}{cat.-specific}
& ZSP~\cite{goodwin2022zero}
& 50 seq. / cat.
& 46.0 & 42.2 & 44.1 \\
& UOP3D~\cite{sommer2024unsupervised}
& 50 seq. / cat.
& 69.2 & 52.4 & 60.8 \\
& Common3D~\cite{sommer2025common3d}
& 50 seq. / cat.
& 75.3 & 56.8 & 66.1 \\
\midrule
\multirow{3}{*}{cat.-agnostic}
& $\text{Ours}^{\text{cat}}_{50}$
& 50 seq. / cat.
& 68.1 & 53.6 & 60.9 \\
& $\text{Ours}^{\text{cat}}_{\text{full}}$
& all seq. / cat.
& \underline{81.6} & \underline{63.1} & \underline{72.4} \\
& Ours
& ${\sim}$160k seq.
& \textbf{85.9} & \textbf{70.0} & \textbf{77.9} \\
\bottomrule
\end{tabular}
\caption{\textbf{Effect of data scale.}
Acc@$30^\circ$ on the 7-category PASCAL3D+ and 20-category ObjectNet3D subsets used by category-specific baselines.
Note that these category subsets differ from the full-dataset evaluation in Table~\ref{tab:zeroshot_combined}, which accounts for the different PASCAL3D+ scores.}
\label{tab:category_transfer}
\end{table}

\subsection{Transfer Across Benchmarks}
\label{sec:exp_zeroshot}

Table~\ref{tab:zeroshot_combined} compares our method to OrientAnything V1/V2 and QWEN3-VL across five benchmarks.
We train once on UCO3D and evaluate on each benchmark without dataset-specific finetuning.
Despite using no canonical pose labels and training only on real video, our method achieves the best average performance. The strongest results appear on Objectron and PASCAL3D+, where most categories have distinctive semantic axes that provide clear orientation signals from multi-view consistency.
On REAL275, all methods benefit from limited pose variability in the dataset, but our method shows the largest margin in Acc@$30^\circ$.
On Omni6DPose, our performance falls behind OrientAnything~V2, which we attribute to its large percentage of symmetric objects (see Section~\ref{sec:exp_analysis}).
Figure~\ref{fig:qualitative_main} shows qualitative pose predictions across diverse categories and benchmarks.

\subsection{Effect of Data Scale}
\label{sec:cat_sspose}

Table~\ref{tab:category_transfer} compares our category-agnostic model to self-supervised baselines that learn canonical frames without canonical pose labels but train \emph{separate} models per category (ZSP/UOP3D/Common3D), using category supervision and relatively little per-category multi-view training data ($\leq$50 sequences per category).
To separate method from data effects, we include two controls trained on the same categories: $\text{Ours}^{cat}_{50}$ trains with the same 50-sequence cap, while $\text{Ours}^{cat}_{full}$ trains on all available UCO3D sequences. 

Our full model achieves the best average Acc@$30^\circ$, improving over Common3D on both PASCAL3D+ ($85.9\%$ vs.\ $75.3\%$) and ObjectNet3D ($70.0\%$ vs. $56.8\%$).
Notably, $\text{Ours}^{cat}_{50}$ remains competitive even without a scale advantage, while scaling within the fixed categories yields substantial gains ($\text{Ours}^{cat}_{50}\!\rightarrow\!\text{Ours}^{cat}_{full}$).
The strongest results are obtained with category-agnostic training (Ours), which highlights that data scale is critical to learn a generalizable representation.

\begin{table}[t]
\centering
\small
\setlength\tabcolsep{3.5pt}
\renewcommand{\arraystretch}{1}
\begin{tabular}{cccc cc cc cc}
\toprule
& & & & \multicolumn{2}{c}{\textbf{Objectron}} & \multicolumn{2}{c}{\textbf{PASCAL3D+}} & \multicolumn{2}{c}{\textbf{ImageNet3D}} \\
\cmidrule(lr){5-6} \cmidrule(lr){7-8} \cmidrule(lr){9-10}
Mesh & PCA & Align. & $K$ & Med$\downarrow$ & Acc30$\uparrow$ & Med$\downarrow$ & Acc30$\uparrow$ & Med$\downarrow$ & Acc30$\uparrow$ \\
\midrule
Cube & \ding{52} & \ding{52} & 4 & 15.3 & 69.2 & 15.7 & 79.8 & 25.5 & 55.0 \\
Cube & \ding{56} & \ding{52} & 4 & 17.7 & 65.1 & 17.5 & 76.6 & 27.5 & 53.1 \\
Cube & \ding{52} & \ding{56} & 4 & 32.0 & 49.1 & 25.0 & 55.0 & 33.6 & 45.0 \\
\midrule
Rect.\ prism & \ding{52} & \ding{52} & 4 & 17.0 & 66.8 & 15.4 & 79.5 & 28.5 & 52.0 \\
Sphere & \ding{52} & \ding{52} & 4 & 19.6 & 63.3 & 17.6 & 75.5 & 27.1 & 53.8 \\
Cylinder & \ding{52} & \ding{52} & 4 & 20.9 & 62.9 & 18.4 & 74.1 & 29.4 & 50.1 \\
Sphere & \ding{56} & \ding{52} & 4 & 21.4 & 62.2 & 23.1 & 72.4 & 29.7 & 50.4 \\
\midrule
Cube & \ding{52} & \ding{52} & 2 & 18.2 & 64.4 & 19.6 & 72.5 & 27.0 & 53.9 \\
Cube & \ding{52} & \ding{52} & 1 & 57.3 & 32.5 & 53.0 & 28.0 & 50.9 & 33.9 \\
\midrule
\multicolumn{4}{l}{Mapping: 24 axis conv.} & 15.3 & 69.1 & 15.7 & 79.8 & 25.7 & 54.9 \\
\multicolumn{4}{l}{Mapping: single global} & 22.7 & 55.3 & 15.6 & 77.5 & 31.5 & 49.0 \\
\bottomrule
\end{tabular}
\caption{\textbf{Ablations.} Each row varies one design choice from the full model (first row). We ablate PCA initialization, learned alignment, mesh shape, and the number of training views $K$ per sequence. The last two rows keep the full model and only restrict the evaluation-time convention mapping (Section~\ref{sec:exp_analysis}).}
\label{tab:ablations}
\end{table}

\subsection{Ablations}
\label{sec:ablations}

Table~\ref{tab:ablations} ablates the canonical mesh shape, PCA initialization, the per-sequence alignment step, and the number of training views $K$ per sequence, reporting transfer accuracy on three benchmarks.

\textbf{PCA and learned alignment are complementary.}
PCA initialization without per-sequence alignment (row 3) maintains reasonable performance on ImageNet3D ($45.0\%$ vs.\ $55.0\%$ for the full model), which contains a high proportion of symmetric objects. 
When we stratify by symmetry class, PCA-only actually improves on continuously symmetric objects ($59.1\% \rightarrow 65.0\%$ for $s{=}\infty$) while degrading severely on non-symmetric objects ($53.6\% \rightarrow 41.7\%$ for $s{=}1$).
This reflects the distinct roles of the two components: PCA captures geometric structure, such as elongation, flatness, spatial extent, which is often sufficient to determine orientation for symmetric objects, but cannot resolve semantic axes such as front-vs-back on objects where multiple orientations are geometrically equivalent. 
The learned alignment adds this semantic disambiguation through appearance-based correspondences (see Figure~\ref{fig:supp_alignment} for an example), producing substantial gains on PASCAL3D+ and Objectron ($55.0\% \rightarrow 79.8\%$, $49.1\% \rightarrow 69.2\%$), where most categories have distinctive semantic axes. 
Conversely, removing PCA while keeping learned alignment (row 2) causes smaller but consistent drops of $2$--$4$ points across benchmarks, confirming that the geometric initialization aids convergence but is not the primary driver, and that the canonical frame emerges from the self-supervised alignment through the bottleneck.

\textbf{Mesh shape matters, but the bottleneck principle is robust.} 
Rows 4--7 compare the cube to a rectangular prism, a sphere, and a cylinder.
The results suggest that the primitive must provide enough geometric structure to anchor orientation, while remaining neutral about which axis acquires which semantic meaning; the cube is the only tested primitive that satisfies both.
The rectangular prism provides strong orientation structure, but its unequal side lengths prescribe long, medium, and short axes, encouraging an extent-based convention that does not transfer consistently across categories.
The sphere is axis-neutral but its continuous symmetry provides no orientation anchors, and the cylinder is both axis-biased and continuously ambiguous around its principal axis.
On the heterogeneous ImageNet3D, the sphere even outperforms the prism ($53.8\%$ vs.\ $52.0\%$), suggesting that avoiding a misleading axis prior can be more valuable than stronger geometric structure.
All primitives nevertheless remain reasonably effective, indicating that the geometric bottleneck concept is not tied to a specific shape.

\textbf{Multi-view coverage matters.}
 Reducing the number of training views per sequence from $K{=}4$ to $K{=}2$ causes moderate drops across benchmarks (e.g., $79.8\% \rightarrow 72.5\%$ on PASCAL3D+), while $K{=}1$ leads to severe degradation ($79.8\% \rightarrow 28.0\%$).
With a single view, the per-sequence alignment becomes underconstrained and causes mode collapse as a trivial solution. 
Two views recover much of the performance, but four views provide the multi-view diversity needed for reliable alignment.

\subsection{Analysis: Frame Consistency, Unseen Categories, and Symmetry}
\label{sec:exp_analysis}

\begin{figure*}[t]
\centering
\begin{minipage}[t]{0.4\textwidth}
\vspace{0pt}
\centering
\includegraphics[width=0.8\linewidth]{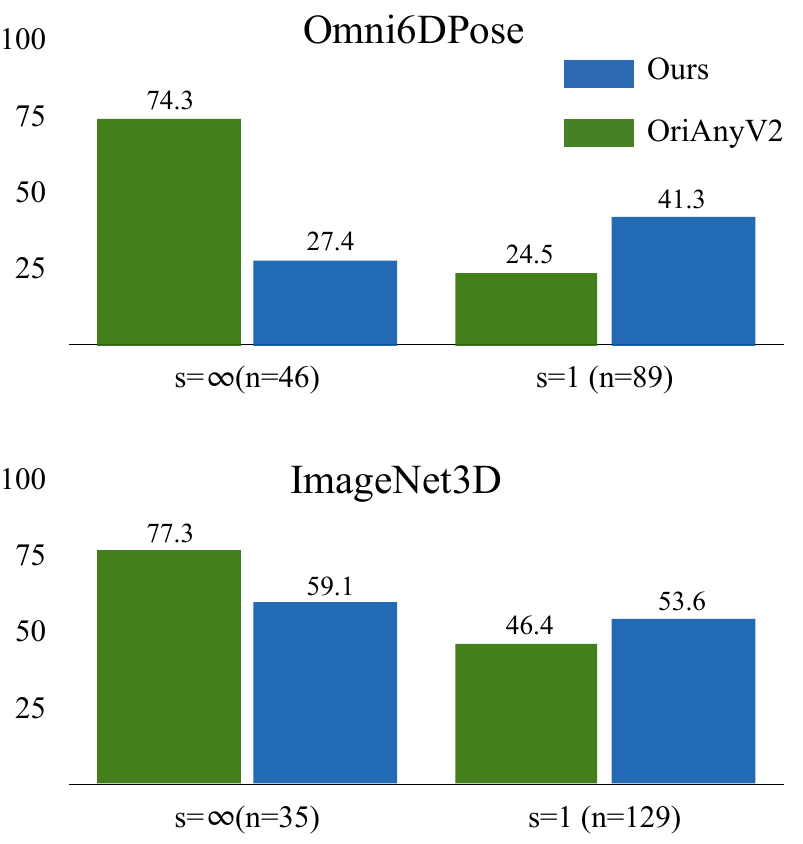}\\
\subcaption{\textbf{Symmetry-stratified accuracy.} Acc$30^\circ$ by symmetry class $s$ on Omni6DPose (top) and ImageNet3D (bottom).}
\label{fig:symmetry_bars}
\end{minipage}\hfill
\begin{minipage}[t]{0.54\textwidth}
\vspace{0pt}
\centering
\includegraphics[width=0.8\linewidth]{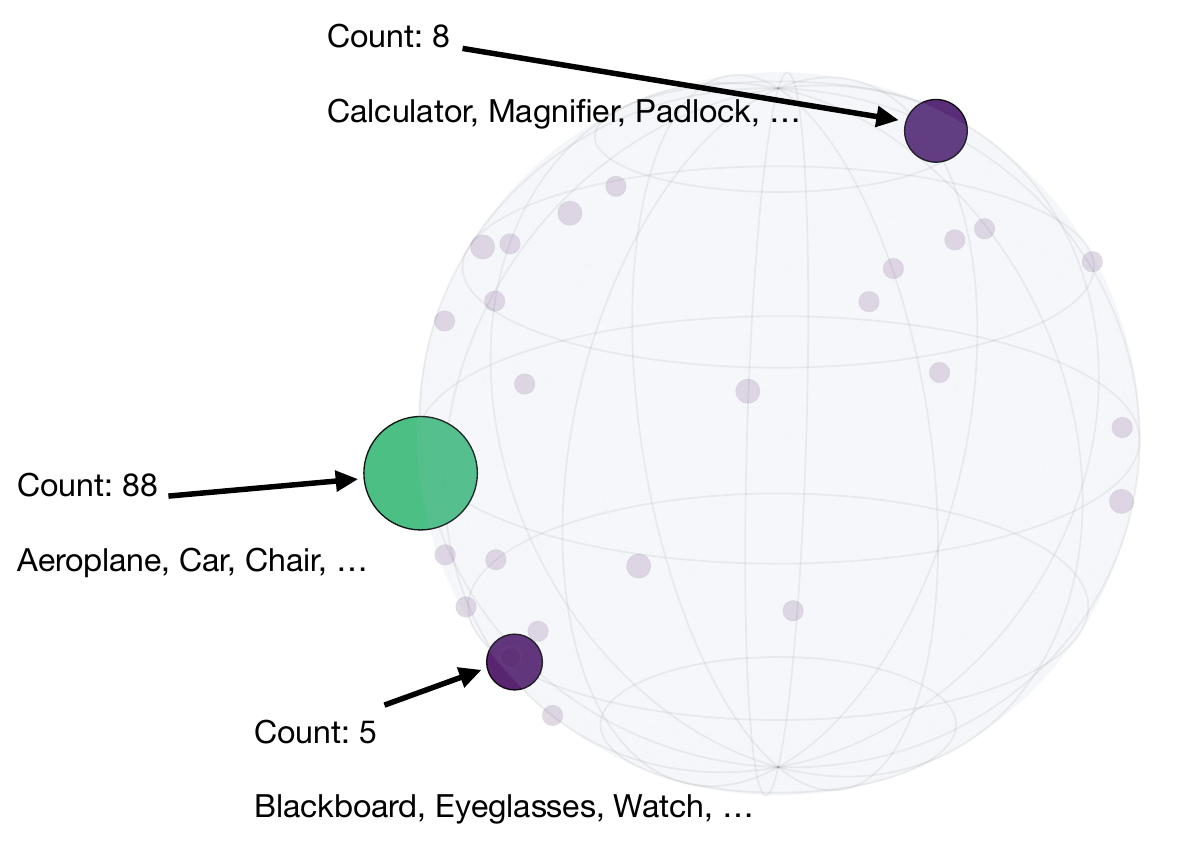}
\subcaption{\textbf{Convention mappings cluster across categories.} Per-category convention mappings (estimated via rotation averaging) concentrate into a few modes, indicating that the learned frame largely agrees with the conventions in ImageNet3D, up to a single rotation.}
\label{fig:canonicalization}
\end{minipage}
\caption{\textbf{Analysis of symmetry and convention mapping.}
Our main performance gap concentrates on symmetry-heavy objects (left), while per-category convention mappings cluster into a small number of modes (right), supporting that the emerged canonical frame is shared across categories.}
\label{fig:analysis_symmetry_canonical}
\end{figure*}

\textbf{Frame consistency.} To assess whether the emerged canonical frame is shared across categories, we estimate per-category convention mappings on ImageNet3D and visualize their agreement.
For each category $c$, we compute the mapping $\hat{\mathbf{A}}_c$ (Appendix~\ref{sec:supp_eval_protocol}) and apply it to a fixed reference vector $\mathbf{v} = (\frac{1}{\sqrt{3}}, \frac{1}{\sqrt{3}}, \frac{1}{\sqrt{3}})^\top$, plotting the resulting direction $\hat{\mathbf{A}}_c \mathbf{v}$ on the unit sphere. 
If the learned frame is consistent, these directions should cluster into a small number of modes. 
Figure~\ref{fig:canonicalization} shows this visualization restricted to non-symmetric categories ($s {=} 1$), removing ambiguity due to rotational symmetry. 
The dominant mode contains 88 of 129 non-symmetric categories, indicating that the learned frame is largely shared across diverse object types. 
The remaining modes are small and largely correspond to categories with geometrically ambiguous structure (e.g., watches, padlocks), where multiple axis assignments are plausible.

\textbf{Reliability of the convention mapping.}
To test how much work the evaluation mapping does, we restrict each per-category mapping from an unrestricted rotation to one of the 24 proper rotational symmetries of the cube, i.e., handedness-preserving permutations and sign changes of the coordinate axes, which cannot absorb fine-grained per-category errors (Table~\ref{tab:ablations}, last two rows).
This changes accuracy by at most 0.1 points: the mapping only selects among discrete axis conventions rather than compensating for inconsistent predictions.
A single global rotation shared by all categories is stricter and performs worse, because a few categories are internally consistent but disagree with the dataset's choice of semantic axes (e.g., books and laptops on Objectron, dining tables on PASCAL3D+).
Per-category results are provided in the appendix.

\textbf{Unseen categories.}
To isolate category-level novelty, we evaluate on the 32 ImageNet3D categories with no exact or synonymous match in the 1,067-category UCO3D vocabulary (4,648 images; e.g., jinrikisha, ferris wheel, hovercraft).
Our method reaches $52.0\%$ Acc@$30^\circ$ on these categories, compared to $55.5\%$ on the remaining 157.
Both groups have a similar symmetry composition ($22/66/13/0\%$ vs.\ $18/69/11/2\%$ for $s{=}\infty/1/2/4$), so the 3.5-point gap reflects category novelty rather than a harder symmetry mix.
While we cannot verify that the pretrained backbone has never encountered these categories, the result supports generalization beyond the training vocabulary.

\textbf{Symmetry.} The emerged frame is most consistent for objects with distinctive semantic axes, where appearance and multi-view constraints provide clear orientation signals. 
For a symmetric object, the reference frame is only defined up to its symmetry group, and any element of that group is a valid choice.
At evaluation, we account for this by minimizing the error over symmetry-equivalent ground-truth rotations.
During training, however, symmetry is not modeled: the alignment (Eq.~\ref{eq:wahba}) selects one valid element of the symmetry group, and the multi-view loss then enforces this single correspondence assignment even though several are equally valid.
The resulting pseudo-labels are geometrically correct but not predictable from appearance, since symmetry-equivalent views are virtually indistinguishable.
Methods supervised with canonical poses do not suffer from this problem, making it significantly easier to learn robust defaults for symmetric objects.
Figure~\ref{fig:symmetry_bars} confirms this: our method falls behind on continuously symmetric objects ($s{=}\infty$) but shows consistently stronger performance on non-symmetric objects ($s{=}1$).
Optimizing over symmetry-equivalent alignments is not a trivial extension: symmetry is only definable relative to a stable frame, which is what training aims to establish, and unconstrained self-supervised symmetry prediction trivially minimizes the loss by declaring an overly broad symmetry group.

\section{Conclusion}
\label{sec:concl}

We showed that a shared canonical frame can emerge from object-centric video of rigid objects without canonical pose labels or category conditioning, by routing all training through a shared geometric bottleneck. 
Across multiple benchmarks, the resulting frame enables competitive pose estimation compared to methods trained with canonical supervision, generalizes to categories outside the training vocabulary, and improves with data volume and category diversity.
Our analysis shows that the emerged frame is largely consistent across categories, and that the evaluation mapping to benchmark conventions reduces to selecting discrete axis conventions.

\textbf{Limitations.}
Our formulation assumes rigid objects: articulated and deformable objects would require a representation beyond a single canonical frame, and for articulated instances (e.g., laptops at different hinge angles) the predicted frame follows whichever part dominates the correspondence evidence.
Chirality is not modeled, so mirrored instances such as left and right shoes share the same canonical frame, consistent with the conventions of the evaluated benchmarks.
Translation is recovered only up to scale, and accuracy drops sharply once more than roughly half of the object is occluded (Appendix~\ref{sec:supp_occlusion}).
Finally, symmetry is handled at evaluation but not during training, which causes our main performance gap on symmetric objects. 
Jointly learning the canonical frame and the symmetry group without canonical pose or symmetry labels remains the principal open problem left by this work.

\begin{ack}
    AK acknowledges support via his Emmy Noether Research Group funded by the German Research Foundation (DFG) under grant number 468670075.
    TF gratefully acknowledges the scientific support and HPC resources provided by the Erlangen National High Performance Computing Center (NHR@FAU) of the Friedrich-Alexander-Universität Erlangen-Nürnberg (FAU) under the BayernKI project \textbf{\textit{v114be}}. BayernKI funding is provided by Bavarian state authorities.
\end{ack}

\bibliographystyle{ieeetr}
\bibliography{main}

\clearpage

\appendix
\section{Overview}
\label{sec:overview}

This supplement provides additional details and analyses referenced in the main paper:
\begin{itemize}[leftmargin=0.5cm, itemsep=2pt]
    \item Implementation details and training hyperparameters (\ref{sec:supp_impl})
    \item Evaluation protocol, including symmetry-aware metrics and convention mapping (\ref{sec:supp_eval_protocol})
    \item Visualization of the learned per-sequence alignment (\ref{sec:supp_alignment})
    \item Reliability of the convention mapping, with per-category results (\ref{sec:supp_mapping})
    \item Robustness to occlusion (\ref{sec:supp_occlusion})
    \item Full symmetry-class breakdown (\ref{sec:supp_symmetry_full})
    \item Mesh-resolution runtime trade-off (\ref{sec:supp_mesh_tradeoff})
    \item PASCAL3D+ and ObjectNet3D per-category results (\ref{supp:per_cate})
    \item Qualitative results (\ref{sec:supp_qual})
    \item LLM-derived symmetry labels (\ref{sec:supp_symmetry_labels})
\end{itemize}

\section{Implementation Details}
\label{sec:supp_impl}
 
\subsection{Data and Preprocessing}

We train on the train split of UCO3D~\cite{uco3d}, which provides object-centric video sequences with RGB frames, segmentation masks, SfM camera poses, and sparse point clouds. 
We undistort images using the provided camera parameters.
Per sequence, we apply the dataset's PCA alignment and normalize the object point cloud using a robust bounding box (10th/90th coordinate quantiles) to fit the unit canonical mesh.
For details about the alignment, we refer the reader to the supplemental material of~\cite{uco3d}.
During training, we sample $K{=}4$ frames per sequence using stratified temporal sampling to encourage viewpoint diversity.
 
\subsection{Data Augmentation}
We apply photometric augmentation throughout training: color jitter (brightness $0.4$, contrast $0.4$, saturation $0.2$, hue $0.1$) with probability $0.8$, random grayscale with probability $0.2$, and Gaussian blur with probability $0.4$. 
Images are cropped to the object mask with $10\%$ padding and resized to $256{\times}256$ with aspect-ratio-preserving padding, followed by ImageNet normalization.
After iteration $7,500$ we enable geometric augmentation: the mask-crop padding is randomly sampled from $[0.1,0.25]$, random recropping is applied with probability $0.7$, in-plane rotation up to $30^\circ$ with probability $0.3$, and random patch occlusion (1--2 patches, $5-30\%$ of the crop each) with probability $0.5$.
All geometric transforms are accompanied by consistent updates to camera intrinsics and extrinsics to preserve multi-view consistency.
The staged schedule avoids destabilizing the co-evolutionary alignment loop during early training, where mode collapse could rarely occur.
 
\subsection{Canonical Mesh}
We use a cube mesh of unit scale with $M{=}1016$ vertices and $2028$ triangular faces. 
The mesh is rasterized at $64{\times}64$ resolution using nvdiffrast~\cite{nvdiffrast}, producing per-pixel object coordinates and a visibility mask for the correspondence loss.
For the primitive ablation (Table~\ref{tab:ablations}), the rectangular prism, sphere, and cylinder use approximately $1{,}000$ evenly sampled vertices and watertight triangulations, and are trained with the same protocol as the cube.

\subsection{Network Architecture}
The visual backbone is DINOv3 ViT-L/16~\cite{dinov3} with LoRA~\cite{hu2022lora} (rank 8, $\alpha{=}8$, dropout 0.1) inserted into the fused query/value projections of every transformer block and the remaining backbone parameters are frozen.
Features from layers 6, 14, 18, and 23 are aggregated into a feature pyramid and decoded with a UPerNet-style decoder~\cite{uppernet} (512 channels) to produce a $64{\times}64$ feature map.
A 6-layer transformer decoder (hidden size 512, 8 heads, feed-forward dimension 2048) decodes one query per mesh vertex.
The correspondence head projects both mesh queries and image features to normalized 512-dimensional descriptors and computes vertex-pixel logits via dot product. 
A separate convolutional head predicts the foreground mask.

\subsection{Training}
We train for 15 epochs (${\sim}75$k iterations) on four H100 GPUs with batch size 8 sequences (effective 32) and mixed precision.
We use AdamW with weight decay 0.05 and gradient clipping at norm 1.0. The learning rate is $10^{-4}$ for non-backbone parameters and $10^{-5}$ for LoRA parameters.
The correspondence softmax temperature $\tau$ follows a three-stage cosine schedule: $0.20 \rightarrow 0.17$ over the first $30\%$ of training, $0.17 \rightarrow 0.13$ over the next $50\%$, and $0.13 \rightarrow 0.10$ over the final 20\%.

\subsection{Robust Per-Sequence Alignment}
 The per-sequence rotation is estimated using a robust Wahba solver with RANSAC (100 iterations, 4-point samples, $25^\circ$ inlier threshold).
 
\subsection{Loss}
The total loss is the per-frame averaged correspondence cross-entropy plus the mask loss ($\lambda{=}1.0$).
The mask loss combines BCE and Dice loss with boundary-focused weighting: pixels within a 2-pixel band of the mask boundary receive full weight, while non-boundary pixels are downweighted by a factor of $0.1$. 
Padded and synthetically occluded regions are excluded from the mask loss.

\subsection{Inference}
We use PoseLib~\cite{poselib} for PnP pose recovery with default settings and a maximum reprojection error threshold of 4.0 pixels.

\section{Evaluation Protocol}
\label{sec:supp_eval_protocol}
 
We evaluate category-level orientation by aligning each predicted pose to the target dataset's coordinate convention and reporting geodesic rotation error and threshold accuracies.
 
\textbf{Symmetry-aware rotation error.}
We measure angular error with the geodesic distance on $\mathrm{SO}(3)$:
\begin{equation}
d(\mathbf{R}^{\text{gt}},\mathbf{R}^{\text{pred}})
=
\arccos\!\left(\tfrac{\mathrm{tr}(\mathbf{R}^{\text{gt}\top} \mathbf{R}^{\text{pred}})-1}{2}\right),
\end{equation}
reported in degrees.
To account for rotational symmetries about the upright axis, we use the category-specific symmetry label $s\in\{\infty,1,2,4\}$ (continuous / none / $180^\circ$ / $90^\circ$ ambiguity) and report the minimum error over the symmetry group:
\begin{equation}
d_{\text{sym}}(\mathbf{R}^{\text{gt}},\mathbf{R}^{\text{pred}})
=
\min_{\mathbf{S}\in\mathcal{S}(s)}
d(\mathbf{R}^{\text{gt}},\mathbf{S}\mathbf{R}^{\text{pred}}),
\end{equation}
where $\mathcal{S}(1)=\{\mathbf{I}\}$, $\mathcal{S}(2)=\{\mathbf{R}_y(0),\mathbf{R}_y(\pi)\}$, $\mathcal{S}(4)=\{\mathbf{R}_y(k\frac{\pi}{2})\}_{k=0}^{3}$, and $\mathbf{R}_y(\theta)$ denotes a rotation around the $y$-axis.
For continuous symmetry ($s{=}\infty$), yaw is unrecoverable and we compare only the transformed upright axes, i.e., the angle between $\mathbf{R}^{\text{gt}}\mathbf{y}$ and $\mathbf{R}^{\text{pred}}\mathbf{y}$ with $\mathbf{y}=(0,1,0)^\top$.
 
\textbf{Per-image inference and pose recovery.}
For each test image, the model predicts dense 2D--3D correspondences from pixels to canonical mesh vertices.
We recover the 3D rotation and scale-normalized translation via EPnP~\cite{epnp} with RANSAC using dataset-provided camera intrinsics.
This yields a predicted rotation $\mathbf{R}^{\text{pred}}$ and scale-normalized translation $\mathbf{t}^{\text{pred}}$ expressed in the learned canonical frame.

\textbf{Convention mapping.}
\label{sec:supp_convention_mapping}
Canonical frames are defined only up to a global rotation, and different benchmarks use different coordinate conventions.
We estimate a per-category convention mapping $\hat{\mathbf{A}}_c$ that aligns our predictions to the dataset convention via rotation averaging~\cite{rotaverage}:
\begin{equation}
\label{eq:procrustes_align}
\hat{\mathbf{A}}_c
=
\argmin_{\mathbf{A}\in\mathrm{SO}(3)}
\sum_{i\in c} \, \big\|\mathbf{R}^{\mathrm{gt}}_{i} - \mathbf{A}\,\mathbf{R}^{\mathrm{pred}}_{i}\big\|_F^2,
\end{equation}
i.e., $\hat{\mathbf{A}}_c$ is the chordal $L_2$ mean of the relative rotations $\mathbf{R}^{\mathrm{gt}}_i\mathbf{R}^{\mathrm{pred}\top}_i$,
which admits a closed-form SVD solution with the $\det(\hat{\mathbf{A}}_c){=}{+}1$ constraint.
This mapping is estimated on the training split of each dataset and applied uniformly to all test predictions.
As shown in the main paper (Figure~\ref{fig:canonicalization}), these per-category mappings cluster into a small number of modes, indicating that the learned frame is largely shared across categories.
Section~\ref{sec:supp_mapping} shows that the mapping only selects among discrete axis conventions and does not compensate for fine-grained per-category errors.

\textbf{Qwen3-VL baseline.}
\label{sec:supp_qwen}
For the QWEN3-VL~\cite{qwen3} baseline, we prompt the model with the input image and request an orientation estimate as rotation angles.
The prompt was taken from the official GitHub repository of the QWEN Team.
We additionally provided the category label in the prompt and bounding box location in the output during our experiments, since we found that this yielded noticeably better results and is likely due to the training paradigm of the model. 
The prompt template is:
 
\begin{quote}\small
    Find the {category} in this image. 
    Provide its 3D bounding box. 
    The output format required is JSON: 
    '`[{{"bbox\_3d":[x\_center, y\_center, z\_center, x\_size, y\_size, z\_size, roll, pitch, yaw]}}]`.'
\end{quote}

\section{Visualization of the Per-Sequence Alignment}
\label{sec:supp_alignment}

\begin{figure}[h]
\centering
\includegraphics[width=\linewidth]{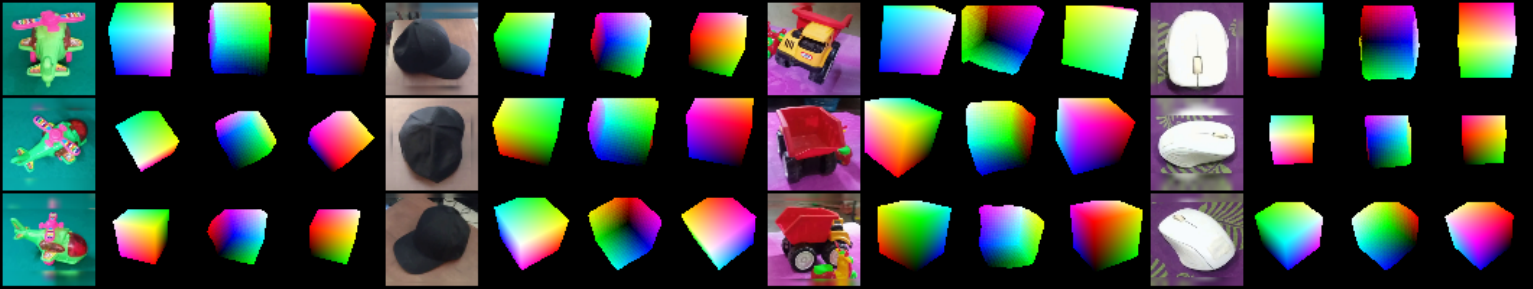}
\caption{\textbf{From PCA initialization to learned alignment.}
Four UCO3D training sequences (helicopter, cap, truck, mouse), each shown for three frames (rows).
Columns per sequence: input frame, coordinate map under the PCA initialization, correspondences $\hat{\mathbf{X}}^c$ predicted by the trained network, and coordinate map under the learned per-sequence alignment $\hat{\mathbf{R}}$ (final pseudo-label).
Colors encode canonical 3D coordinates.
Under PCA, the red corner marks different object parts across sequences (e.g., front-right-bottom for the helicopter, back-left-bottom for the cap). After alignment, it marks the back-right-top of all four objects.}
\label{fig:supp_alignment}
\end{figure}

Figure~\ref{fig:supp_alignment} visualizes the pseudo-labels of the per-sequence alignment (Section~\ref{sec:alignment}) for four UCO3D training sequences.
For each sequence, we render the canonical coordinate map under the PCA alignment provided with the dataset, which serves as initialization, and under the learned alignment $\hat{\mathbf{R}}$, which yields the final pseudo-labels.
We also show the correspondences predicted by the trained network.
The learned alignment replaces the extent-based PCA axes with a rotation estimated from the network's predictions (Eq.~\ref{eq:wahba}), and, as a single rotation per sequence, is applied consistently to all frames.

The colors make the effect on semantic orientation visible.
Under the PCA initialization, the red corner of the cube marks a different part of each object: front-right-bottom for the helicopter and the mouse, but back-left-bottom for the cap and the truck.
After alignment, it marks the back-right-top of all four objects.
The learned alignment thus resolves the front-back and left-right ambiguities left by PCA and places these objects, from four different categories, in a common semantic frame.

\section{Reliability of the Convention Mapping}
\label{sec:supp_mapping}

All results use the per-category mapping $\hat{\mathbf{A}}_c$ of Eq.~\eqref{eq:procrustes_align}.
An unrestricted rotation could, in principle, absorb category-specific prediction errors.
To test this, we compare three variants, all estimated on the training split and never on test data:
\begin{itemize}[leftmargin=0.5cm, itemsep=2pt]
    \item \textbf{Unrestricted (per category):} $\hat{\mathbf{A}}_c \in \mathrm{SO}(3)$ as in Eq.~\eqref{eq:procrustes_align}. This is the default used throughout the paper.
    \item \textbf{24 axis conventions (per category):} $\hat{\mathbf{A}}_c$ is restricted to the 24 proper rotational symmetries of the cube, i.e., handedness-preserving permutations and sign changes of the coordinate axes. We select the element that minimizes the objective of Eq.~\eqref{eq:procrustes_align}. 
    \item \textbf{Single global rotation:} one $\hat{\mathbf{A}}$ shared by all categories of a dataset, obtained by pooling all training samples in Eq.~\eqref{eq:procrustes_align}.
\end{itemize}
The aggregate results are reported in the main paper (Table~\ref{tab:ablations}, last two rows).
The restricted variant cannot correct fine-grained errors, yet it changes Acc@$30^\circ$ by at most $0.1$ points.
The mapping therefore only selects among discrete axis conventions.

Table~\ref{tab:supp_mapping_percat} shows the per-category results on Objectron and PASCAL3D+.
The single global rotation degrades on only a few categories.
These categories are predicted consistently in our frame, but their semantic axes are assigned differently by the dataset convention (books and laptops on Objectron, dining tables on PASCAL3D+).

\begin{table}[h]
\centering
\small
\setlength\tabcolsep{3pt}
\resizebox{\linewidth}{!}{
\begin{tabular}{l ccccccccc | c}
\toprule
\multicolumn{11}{c}{\textbf{Objectron}} \\
Mapping & bike & book & bottle & camera & cereal box & chair & cup & laptop & shoe & Mean \\
\midrule
Unrestricted & 98.1 & 51.3 & 45.3 & 75.5 & 66.3 & 87.0 & 37.7 & 80.5 & 81.7 & 69.2 \\
24 axis conv. & 98.1 & 51.3 & 45.3 & 75.5 & 66.3 & 87.0 & 37.7 & 80.5 & 81.0 & 69.1 \\
Single global & 98.1 & \hphantom{0}1.5 & 45.3 & 75.5 & 66.3 & 87.0 & 37.7 & \hphantom{0}6.4 & 81.0 & 55.3 \\
\bottomrule
\end{tabular}}

\vspace{6pt}
\resizebox{\linewidth}{!}{
\begin{tabular}{l cccccccccccc | c}
\toprule
\multicolumn{14}{c}{\textbf{PASCAL3D+}} \\
Mapping & aero & bicycle & boat & bottle & bus & car & chair & table & mbike & sofa & train & tv & Mean \\
\midrule
Unrestricted & 79.1 & 80.0 & 82.3 &	66.8 &  97.7 & 96.7 & 95.6 & 47.7 & 81.5 & 84.6 & 80.9 & 65.1 & 79.8 \\
24 axis conv. & 79.1 & 80.0 & 82.3 &	66.8 &  97.7 & 96.7 & 95.6 & 47.7 & 81.5 & 84.6 & 80.9 & 65.1 & 79.8 \\
Single global & 79.1 & 80.0 & 82.3 &	66.8 &  97.7 & 96.7 & 95.6 & 20.7 & 81.5 & 84.6 & 80.9 & 65.1 & 77.5 \\
\bottomrule
\end{tabular}}
\caption{\textbf{Per-category convention-mapping results} (Acc@$30^\circ$).
Restricting the per-category mapping to the 24 axis conventions of the cube leaves accuracy virtually unchanged, so the mapping does not absorb per-category errors.
A single global rotation affects only a few categories whose semantic axes are defined differently by the dataset.}
\label{tab:supp_mapping_percat}
\end{table}

\section{Robustness to Occlusion}
\label{sec:supp_occlusion}
 
UCO3D contains little natural occlusion, so we apply random patch occlusion to the input images during training.
We evaluate robustness at inference in two settings: realistic occluders from OccludedPASCAL3D+, and a controlled occluder whose size we vary.
In both settings, all methods use the convention mapping and evaluation protocol of the main paper.
 
\subsection{OccludedPASCAL3D+}
\label{sec:supp_occluded_pascal}
 
OccludedPASCAL3D+~\cite{occludedpascal} places occluders on PASCAL3D+ images at three levels: L1 (20--40\%), L2 (40--60\%), and L3 (60--80\% of the object occluded).
L0 refers to the original, unoccluded images.
Table~\ref{tab:supp_occluded_pascal} shows the results.
Our method is strongest at L0 and L1, and within 0.5 points of the best result at L2, but falls behind at L3.
Resolution and training data both differ between methods, so this comparison is not controlled for either.
OrientAnything~V2 uses $518{\times}518$ inputs, while ours are $256{\times}256$.
The daggered models were trained on ImageNet3D, which overlaps with PASCAL3D+.
We do not claim that these factors fully account for the L3 gap, but they are relevant for interpreting it.
We attribute our robustness at moderate occlusion to patch-occlusion augmentation during training and RANSAC at inference, which rejects correspondences predicted on occluded pixels.
 
\begin{table}[h]
\centering
\small
\setlength\tabcolsep{6pt}
\begin{tabular}{l cccc}
\toprule
Method & L0 & L1 & L2 & L3 \\
\midrule
OriAny.V1~\cite{orientanything} & 71.0 & 66.9 & 61.3 & 48.0 \\
OriAny.V1$^\dagger$~\cite{orientanything} & \textcolor{lightgray}{\underline{78.4}} & \textcolor{lightgray}{\underline{78.0}} & \textcolor{lightgray}{\textbf{75.9}} & \textcolor{lightgray}{64.5} \\
OriAny.V2$^\dagger$~\cite{orientanythingv2} & \textcolor{lightgray}{76.5} & \textcolor{lightgray}{75.0} & \textcolor{lightgray}{74.0} & \textcolor{lightgray}{\textbf{70.6}} \\
\midrule
Ours & \textbf{79.8} & \textbf{79.5} & \underline{75.4} & \underline{63.0} \\
\bottomrule
\end{tabular}
\caption{\textbf{Occlusion robustness on OccludedPASCAL3D+} (Acc@$30^\circ$).
L0 denotes the unoccluded images; L1--L3 correspond to 20--40\%, 40--60\%, and 60--80\% occlusion.
Daggered methods were trained on ImageNet3D (overlapping with PASCAL3D+, gray).
Each method is evaluated at its standard input resolution.}
\label{tab:supp_occluded_pascal}
\end{table}
 
\subsection{Controlled Occlusion Sweep}
\label{sec:supp_occluder_sweep}
 
To measure how accuracy degrades with the occluded fraction rather than in coarse bins, we place a single black rectangular occluder inside the ground-truth 2D bounding box of each PASCAL3D+ object and increase the fraction of the box it covers from $0\%$ to $80\%$ in steps of $10\%$.
The model, convention mapping, and evaluation protocol match the main paper.

\begin{figure}[h]
\centering
\includegraphics[width=\linewidth]{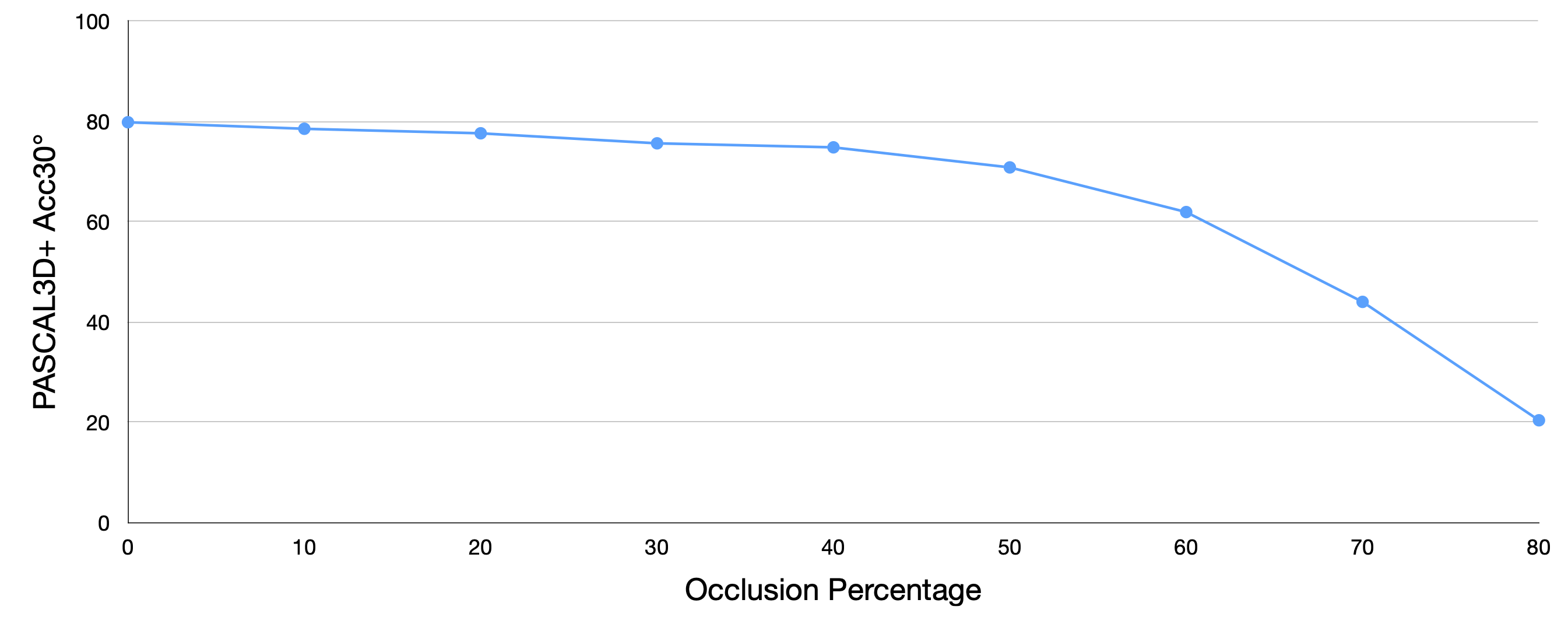}
\caption{\textbf{Controlled occlusion sweep on PASCAL3D+.} A rectangular occluder covers an increasing fraction of the object bounding box. Accuracy degrades gracefully up to ${\sim}40\%$ coverage and drops sharply beyond $50\%$, once most discriminative appearance is removed.}
\label{fig:supp_occluder_sweep}
\end{figure}

Figure~\ref{fig:supp_occluder_sweep} shows the results.
Up to ${\sim}40\%$ coverage, accuracy degrades gracefully ($79.8\% \rightarrow 74.8\%$).
Beyond $50\%$, it drops sharply ($70.8\%$ at $50\%$, $44.0\%$ at $70\%$, $20.4\%$ at $80\%$), once the occluder removes most of the object's discriminative appearance.
This boundary is the occlusion limitation stated in Section~\ref{sec:concl}.

\section{Full Symmetry Breakdown}
\label{sec:supp_symmetry_full}
 
\begin{figure}[t]
    \centering
    \includegraphics[width=\linewidth]{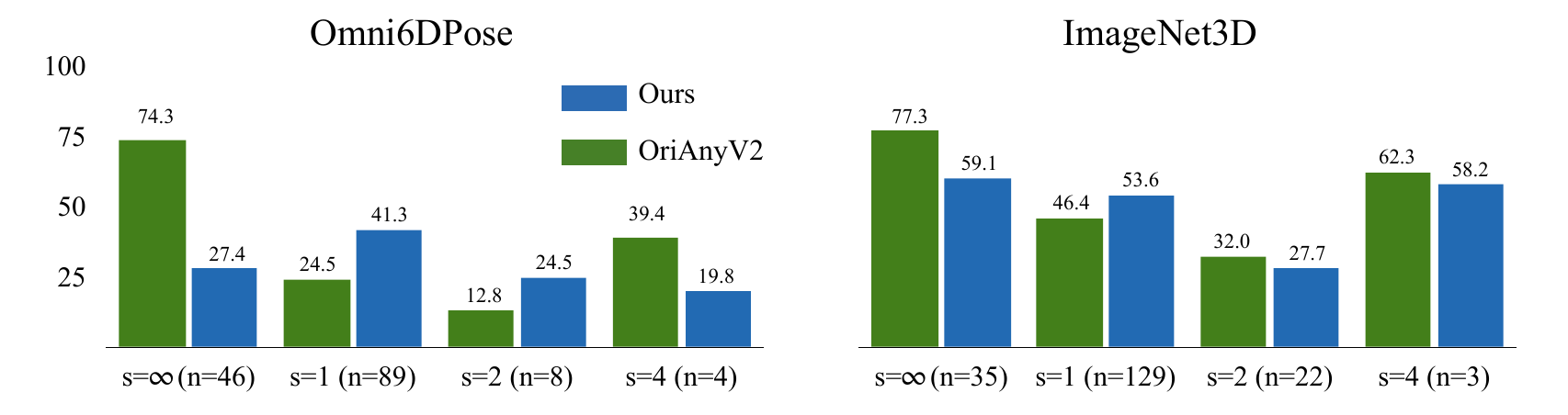}
    \caption{\textbf{Accuracy by symmetry class (full breakdown).}
    Symmetry-aware Acc@$30^\circ$ stratified by $s \in \{\infty,1,2,4\}$ for Omni6DPose and ImageNet3D.
    Bars report accuracy, group sizes are shown alongside each class.}
    \label{fig:supp_symmetry_full}
\end{figure}

The main paper (Figure \ref{fig:analysis_symmetry_canonical}) reports symmetry-stratified accuracy for the two most common classes ($s{=}\infty$ and $s{=}1$).
Figure~\ref{fig:supp_symmetry_full} provides the complete breakdown across all symmetry classes $s \in \{\infty, 1, 2, 4\}$ on ImageNet3D and Omni6DPose.
Two-fold and four-fold symmetries are rare in both datasets, and performance differences between methods are less systematic for these classes.
On two-fold symmetric categories in ImageNet3D ($n{=}22$), our method falls slightly behind OrientAnything~V2, which was trained on ImageNet3D and has seen all categories during training.

\begin{figure}[h]
    \centering
    \includegraphics[width=0.5\linewidth]{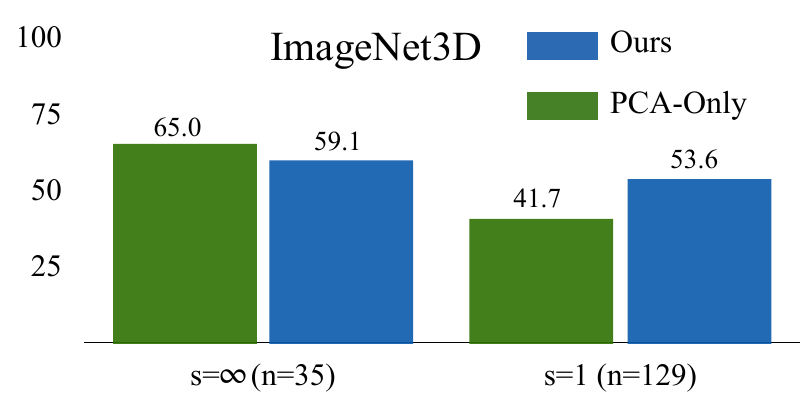}
    \caption{\textbf{Accuracy with and without per-sequence alignment.} Symmetry-aware Acc@$30^\circ$ stratified by $s \in \{\infty,1\}$ for ImageNet3D. Bars report accuracy, group sizes are shown alongside each class.}
    \label{fig:pcavsfull}
\end{figure}

Figure~\ref{fig:pcavsfull} additionally compares our full method to the PCA-only ablation without per-sequence alignment.
For continuously symmetric categories, the PCA-only model outperforms the full method, showing that PCA is an effective gravity prior.
For non-symmetric categories, the per-sequence alignment is necessary to disambiguate semantic axes.
 
\section{Mesh Resolution vs.\ Runtime Trade-off}
\label{sec:supp_mesh_tradeoff}
 
\begin{figure}[t]
    \centering
    \includegraphics[width=\linewidth]{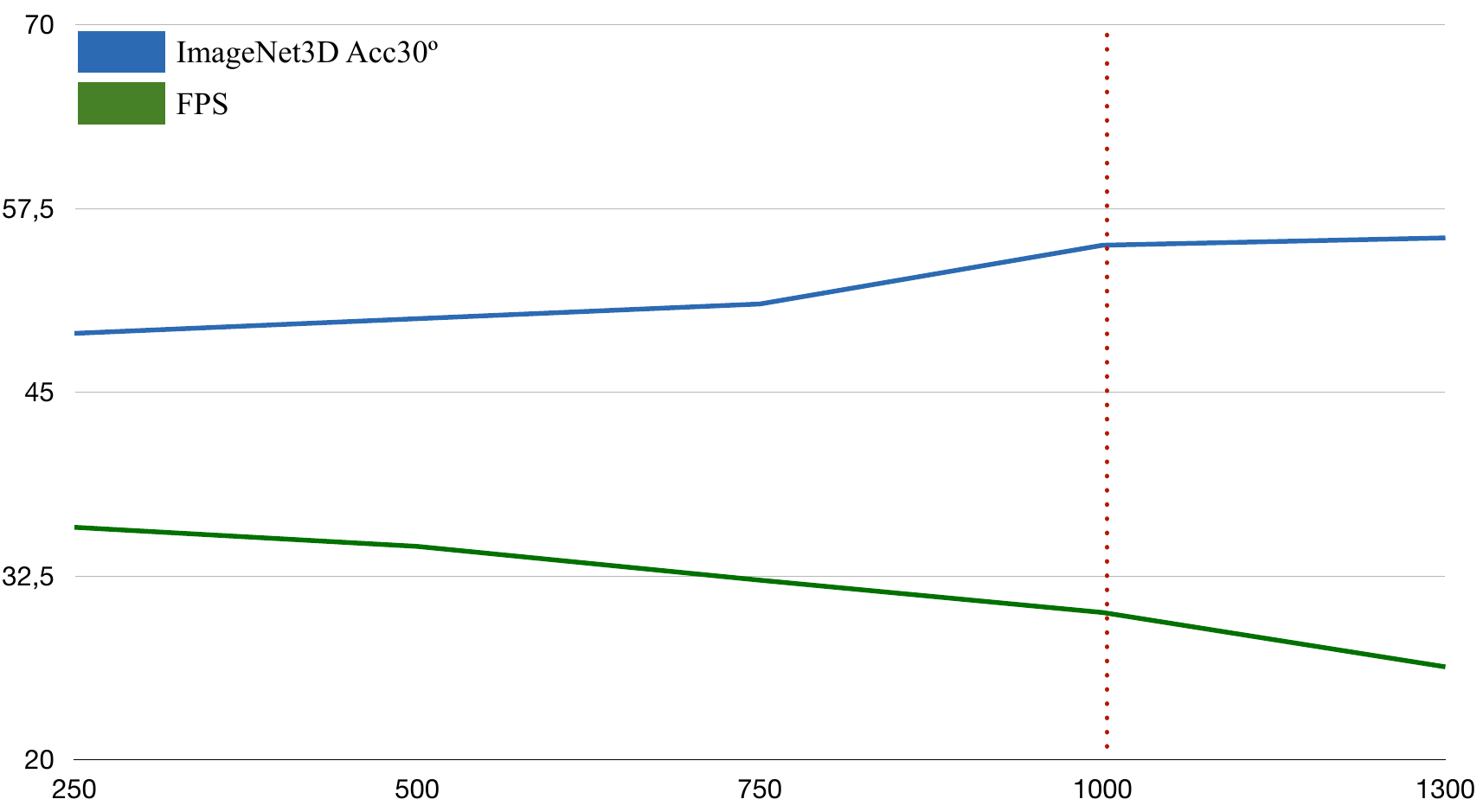}
    \caption{\textbf{Accuracy--runtime trade-off.}
    ImageNet3D Acc@$30^\circ$ (blue) and inference speed in FPS (green) as a function of mesh vertices $M$.
    Red dotted line: our default ($M{=}1000$).}
    \label{fig:supp_acc_fps_tradeoff}
\end{figure}

Our correspondence target is a fixed primitive mesh with $M$ surface vertices.
Increasing $M$ can improve correspondence granularity and downstream pose accuracy, but increases compute and memory through the vertex-query decoding and matching steps.
Figure~\ref{fig:supp_acc_fps_tradeoff} quantifies this trade-off by sweeping $M$ and reporting ImageNet3D Acc@$30^\circ$ together with inference throughput (FPS).
Accuracy improves with mesh resolution and begins to saturate, while FPS decreases monotonically.
Higher vertex counts would likely benefit more if we increased the model's output resolution beyond the $64{\times}64$ grid, but this would substantially impact runtime.
We select $M{=}1000$ as a practical operating point that captures most of the accuracy gains while maintaining favorable throughput.
Runtime was measured on a NVIDIA A6000 GPU.

\section{Per-Category Results on ObjectNet3D and PASCAL3D+}
\label{supp:per_cate}

Figure~\ref{fig:objectnet_pascal_side_by_side} reports per-category $Acc@30^\circ$ on the 20-category ObjectNet3D split and the 7-category PASCAL3D+ split used by Common3D~\cite{sommer2025common3d}.
Baselines train separate models per category on approximately 50 sequences each.
On ObjectNet3D, our full model achieves the highest average ($70.0\%$) and leads on 15 of 20 categories. 
The largest gains appear on categories with distinctive geometric and semantic structure, such as bicycle ($89.1\%$), chair ($98.8\%$), and couch ($98.7\%$), where the combination of geometric and semantic priors provides strong orientation signals.
Categories where our method trails tend to be small, geometrically ambiguous objects where category-specific training can exploit shape priors that a shared model cannot. 
The scaling trend is consistent across most categories: $\text{Ours}^{\text{cat}}_{50} \rightarrow \text{Ours}^{\text{cat}}_{\text{full}} \rightarrow \text{Ours}$ improves performance progressively, with the full category-agnostic model often substantially outperforming the data-matched $\text{Ours}^{\text{cat}}_{50}$, indicating that cross-category training transfers useful structure even to categories seen during training.

On PASCAL3D+, the pattern is similar: our full model leads on 6 of 7 categories and achieves $85.9\%$ average. 
The largest gains over Common3D appear on chair ($+27.8$), bus ($+14.8$), and TV monitor ($+14.0$).
TV monitor remains the hardest category ($65.1\%$), as its near-symmetric geometry limits the orientation signal available from multi-view consistency.

\begin{figure*}[t]
\centering

\begin{minipage}[t]{0.6\textwidth}
\centering
\scriptsize
\setlength\tabcolsep{3pt}
\renewcommand{\arraystretch}{1.12}
\resizebox{\linewidth}{!}{
\begin{tabular}{l|rrrrrrrrrr|r}
\toprule
& \multicolumn{10}{c|}{\textbf{ObjectNet3D (20 categories)}} & \textbf{AVG} \\
\midrule
& \Backpack & \Bench & \faIcon{bicycle} & \faIcon{bus} & \faIcon{car} & \faIcon{phone} & \faIcon{chair} & \faIcon{couch} & cup & \Hairdryer & \\
\midrule
ZSP~\cite{goodwin2022zero}
& 23.1 & 50.8 & 58.6 & 30.5 & 60.3 & 46.4 & 36.8 & 55.5 & 33.0 & \underline{21.7} & 42.2 \\
UOP3D~\cite{sommer2024unsupervised}
& 18.0 & 62.1 & 57.8 & 78.3 & 98.1 & \textbf{54.6} & 52.2 & 76.6 & 38.2 & 14.1 & 52.4 \\
Common3D~\cite{sommer2025common3d}
& 25.5 & \underline{77.0} & 65.7 & 81.1 & \underline{99.0} & \underline{51.8} & 65.9 & 79.0 & 44.7 & 19.4 & 56.8 \\
\midrule
$\text{Ours}^{cat}_{50}$
& 24.7 & 68.2 & 76.0 & 72.9 & 89.7 & 20.0 & 54.6 & 86.8 & 38.2 & \hphantom{0}9.0 & 53.6 \\
$\text{Ours}^{cat}_{full}$
& \underline{41.5} & 74.4 & \underline{83.6} & \underline{93.7} & 98.2 & 36.8 & \underline{94.2} & \underline{91.6} & \underline{66.0} & \hphantom{0}4.9 & \underline{63.1}\\
Ours
& \textbf{45.5} & \textbf{86.7} & \textbf{89.1} & \textbf{95.0} & \textbf{99.1} & 45.1 & \textbf{98.8} & \textbf{98.7} & \textbf{68.5} & \textbf{30.0} & \textbf{70.0} \\
\midrule
& \faIcon{keyboard} & \faIcon{laptop} & \Microwave & \faIcon{motorcycle} & \Mouse & \Remote & \faIcon{suitcase} & \Toaster & \Toilet & \faIcon{tv} & \\
\midrule
ZSP~\cite{goodwin2022zero}
& \textbf{46.8} & \underline{60.5} & 50.5 & 50.3 & 28.8 & 41.6 & 25.8 & 28.8 & 56.3 & 37.9 & 42.2 \\
UOP3D~\cite{sommer2024unsupervised}
& 26.9 & 53.3 & \textbf{80.3} & 69.0 & 44.7 & \textbf{54.4} & 15.5 & \underline{60.6} & 39.6 & 53.2 & 52.4 \\
Common3D~\cite{sommer2025common3d}
& 34.5 & 54.2 & \underline{76.1} & 81.0 & 38.5 & \underline{52.4} & 15.7 & 57.3 & 65.0 & 52.4 & 56.8 \\
\midrule
$\text{Ours}^{cat}_{50}$
& 28.9 & 45.4 & 68.3 & 80.6 & 79.9 & 32.2 & 20.1 & 51.3 & 57.3 & \underline{67.9} & 53.6 \\
$\text{Ours}^{cat}_{full}$
& 38.3 & \underline{60.5} & 58.4 & \underline{87.3} & \underline{82.5} & 34.0 & \underline{26.1} & \textbf{63.6} & \underline{81.3} & 44.6 & \underline{63.1} \\
Ours
& \underline{43.2} & \textbf{74.1} & 66.6 & \textbf{88.3} & \textbf{86.3} & 36.9 & \textbf{56.0} & 36.3 & \textbf{84.9} & \textbf{71.6} & \textbf{70.0} \\
\bottomrule
\end{tabular}}
\subcaption{\textbf{ObjectNet3D.} Acc@$30^\circ$(\%) on the 20-category Common3D\cite{sommer2025common3d} split.}
\label{fig:objectnet3d_side}
\end{minipage}
\hfill
\begin{minipage}[t]{0.39\textwidth}
\centering
\scriptsize
\setlength\tabcolsep{3pt}
\renewcommand{\arraystretch}{1.12}
\resizebox{\linewidth}{!}{
\begin{tabular}{l|ccccccc|c}
\toprule
& \multicolumn{7}{c|}{\textbf{PASCAL3D+ (7 categories)}} & \textbf{AVG} \\
\midrule
  & \faIcon{bicycle} & \faIcon{bus} & \faIcon{car} & \faIcon{chair} & \faIcon{couch} & \faIcon{motorcycle} & \multicolumn{1}{l|}{\faIcon{tv}} & \\
\midrule
ZSP  & 61.7 & 21.4 & 61.6 & 42.6 & 52.9 & 43.1 & 39.0 & 46.0 \\
UOP3D  & 58.4 & 79.3 & \underline{98.2} & 51.9 & 76.6 & 67.0 & 53.1 & 69.2 \\
Common3D  & 67.2 & 82.9 & \textbf{99.3} & 67.8 & 80.9 & 78.1 & 51.1 & 75.3 \\
\midrule
$\text{Ours}^{cat}_{50}$  & 67.4 & 68.8 & 81.3 & 50.7 & 79.6 & 69.0 & 59.6 & 68.1 \\
$\text{Ours}^{cat}_{full}$ & \underline{74.4} & \underline{93.8} & 90.4 & \underline{89.9} & \underline{82.9} & \underline{78.7} & \underline{61.0} & \underline{81.6} \\
\midrule
Ours  & \textbf{80.0} & \textbf{97.7} & 96.7 & \textbf{95.6} & \textbf{84.6} & \textbf{81.5} & \textbf{65.1} & \textbf{85.9} \\
\bottomrule
\end{tabular}}
\subcaption{\textbf{PASCAL3D+.} Acc@$30^\circ$ (\%) on the 7-category protocol of Common3D~\cite{sommer2025common3d}.}
\label{fig:pascal3d_side}
\end{minipage}
\caption{
\textbf{Category-level pose estimation results.}
Left: ObjectNet3D results on the 20-category Common3D split.
Right: PASCAL3D+ results on 7 categories.
Baselines train separate models per category on roughly 50 sequences.
Our fairness controls show scaling effects: $\text{Ours}^{cat}_{50}$ trains on the same target categories with a 50-sequence-per-category cap, $\text{Ours}^{cat}_{full}$ trains on the same target categories using all available UCO3D sequences, and \textbf{Ours} trains category-agnostically on all UCO3D.
}
\label{fig:objectnet_pascal_side_by_side}
\end{figure*}

\section{Qualitative Results}
\label{sec:supp_qual}
 
\subsection{Pose Predictions Across Benchmarks}
 
\begin{figure}[h]
    \centering
    \includegraphics[width=\linewidth]{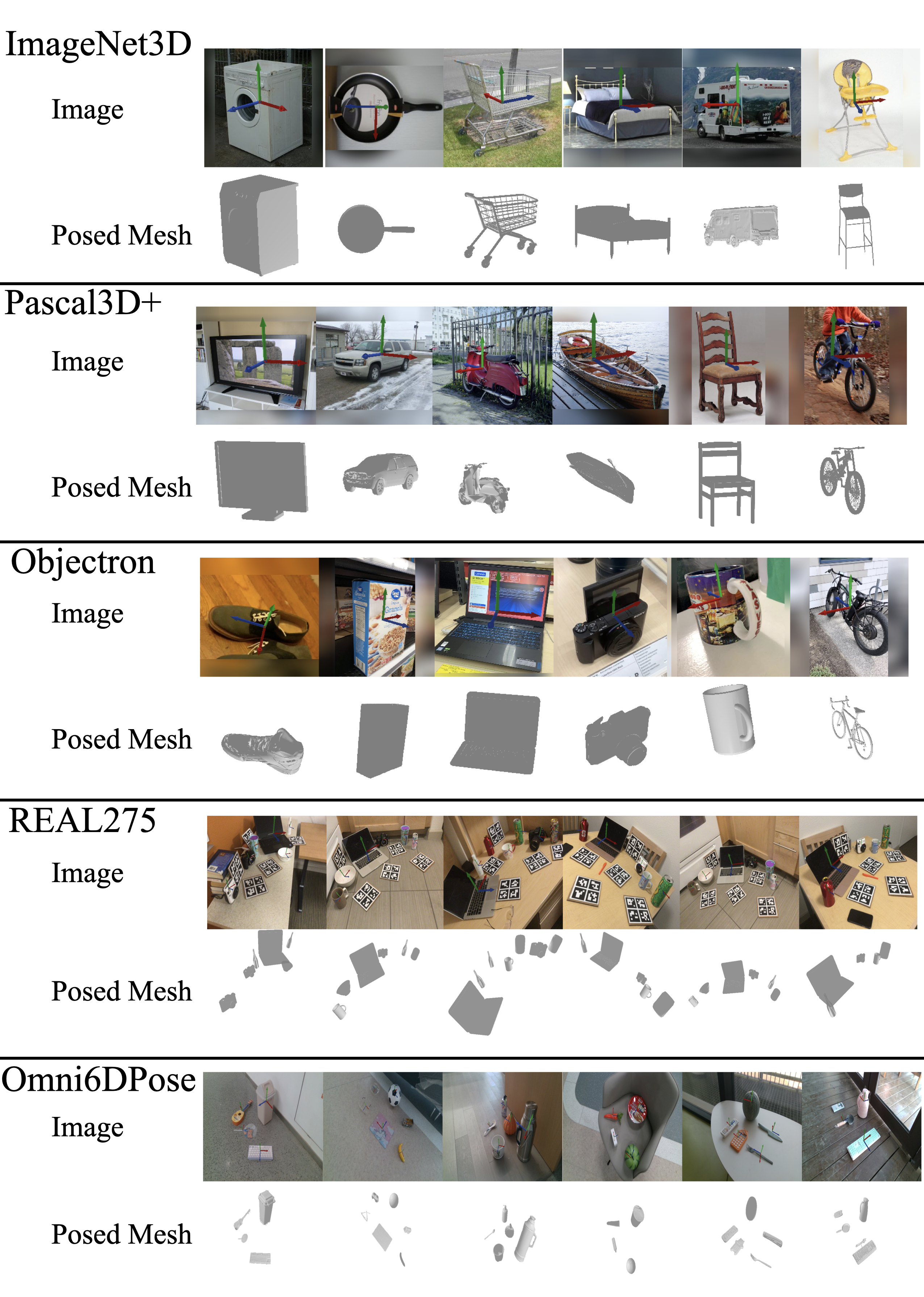}
    \caption{\textbf{Qualitative pose predictions across benchmarks.}
    For each dataset: input images with predicted canonical axes (top) and ShapeNet meshes rendered under the predicted pose (bottom).
    Blue axis indicates the predicted ``front'' direction.}
    \label{fig:supp_qualitative}
\end{figure}

Figure~\ref{fig:supp_qualitative} visualizes predicted poses across the five evaluation benchmarks.
For each dataset we show input images overlaid with predicted canonical axes (top) and a ShapeNet~\cite{chang2015shapenet} mesh rendered under the predicted pose (bottom).
Because RGB-only pose is evaluated in a scale-normalized coordinate system, rendered meshes use a fixed canonical scale and translations are comparable only up to a global scale factor.
The examples illustrate that the learned canonical frame induces consistent axis conventions across categories and datasets.
 
\section{LLM-Derived Symmetry Labels}
\label{sec:supp_symmetry_labels}
 
Several evaluations use symmetry-aware rotation metrics, where the error for a prediction is the minimum over symmetry-equivalent ground-truth rotations (following~\cite{orientanythingv2}).
This requires assigning each category a discrete yaw symmetry class.
We derive these labels automatically using a large language model to avoid manual annotation.
These labels are used only at evaluation time and are applied uniformly across all compared methods; they are not used during training.
 
\subsection{Symmetry Classes}
 
We assign each category to one of four classes:
\begin{enumerate}[itemsep=2pt]
    \item \textbf{Class} $\infty$ {(continuous)}: rotationally symmetric about the vertical axis; yaw is unrecoverable.
    \item \textbf{Class 1 (asymmetric)}: a unique identifiable front; full yaw determination is possible.
    \item \textbf{Class 2 ($180^\circ$ ambiguity)}: front and back are indistinguishable but side views differ.
    \item \textbf{Class 4 ($90^\circ$ ambiguity)}: all four cardinal yaw orientations are indistinguishable.
\end{enumerate}
 
\subsection{LLM and Prompt Details}
 
We generated symmetry labels using ChatGPT (GPT-5.2), queried in February 2026 with default decoding settings.
We used a single fixed prompt and queried the model with the full list of categories from all considered datasets.
We did not iterate on the prompt or revise individual labels based on downstream results.
To simplify the prompt, we replaced symmetry class name $\infty$ with $0$ in the prompt.
The complete category-to-symmetry mapping is provided as a CSV file with the released code.
 
\textbf{Prompt:}
\begin{quote}\small
I have a list of object categories that I need labeled by their object pose symmetry class.
Each object category is assigned a discrete symmetry class based on the degree of pose ambiguity an observer would face when estimating the object's yaw from a single view:
 
- Class 0 (Continuous): The object is rotationally symmetric about its vertical axis, making yaw unrecoverable. Applies to cylindrical (bottle, can), spherical (ball, orange), round flat (plate, pizza), and generally blob-shaped objects (pineapple, hamburger, hat) --- even if minor surface details exist.
- Class 1 (Asymmetric): The object has a unique, identifiable front face, allowing full pose determination. This is the default class and covers most man-made objects including vehicles, electronics, furniture with a clear front, labeled containers, tools, and figures.
- Class 2 (180° ambiguity): Front and back views are indistinguishable, but side views differ. Applies to bilaterally symmetric elongated objects (barbell, pen, sausage) and objects symmetric through their dividing plane (bench, pillow, radiator).
- Class 4 (90° ambiguity): All four cardinal views are identical. Applies to objects with square cross-sectional symmetry (table, box, dice, stool).
 
Please label the following categories with their symmetry class (0, 1, 2, or 4) and output the result as a CSV with columns category,symmetry\_class:
\end{quote}
 
\subsection{Sample Annotations}
\begin{verbatim}
category,symmetry_class
aeroplane,1
air_hammer,1
airship,1
ambulance,1
ashtray,0
ax,1
backpack,1
ball,0
banana,2
barbell,2
baseball_bat,0
basket,0
\end{verbatim}

\end{document}